\documentclass{article}

\usepackage{microtype}
\usepackage{graphicx}
\usepackage{booktabs} 
\usepackage{xspace}
\usepackage[frozencache,cachedir=.]{minted}
\usepackage{subcaption}
\usepackage{enumitem}
\usepackage{tabularx}
\usepackage{listings}
\usepackage[]{algorithm2e}
\usepackage{amsmath}
\usepackage{url}

\usepackage{breakurl}
\usepackage[breaklinks]{hyperref}

\usepackage[small,compact]{titlesec}
\usepackage[subtle]{savetrees}

\usepackage{hyperref}

\usepackage[accepted]{mlsys2020}

\newcommand\Small{\fontsize{8}{8.2}\selectfont}
\newcommand*\LSTfont{\Small\ttfamily\SetTracking{encoding=*}{-60}\lsstyle}
\newcommand{\code}[1]{\texttt{\small#1}}

\newtoggle{rcolors} \togglefalse{rcolors}
\newcommand{\colora}[1]{\iftoggle{rcolors}{{\color{red}{#1}}}{#1}}

\DeclareMathOperator*{\argmax}{arg\,max}

\newcommand{\specialcell}[2][c]{%
    \begin{tabular}[#1]{@{}l@{}}#2\end{tabular}}
\newcommand{\minihead}[1]{{\vspace{.45em}\noindent\textbf{#1.} }}
\newcommand{\sn}{\textsc{OMG}\xspace}

\mlsystitlerunning{Model Assertions for Monitoring and Improving ML Models}

\begin{document}

\twocolumn[
\mlsystitle{Model Assertions for Monitoring and Improving ML Models}



\mlsyssetsymbol{equal}{*}

\begin{mlsysauthorlist}
\mlsysauthor{Daniel Kang}{equal,stan}
\mlsysauthor{Deepti Raghavan}{equal,stan}
\mlsysauthor{Peter Bailis}{stan}
\mlsysauthor{Matei Zaharia}{stan}
\end{mlsysauthorlist}

\mlsysaffiliation{stan}{Stanford University}

\mlsyscorrespondingauthor{Daniel Kang}{ddkang@stanford.edu}


\vskip 0.3in

\begin{abstract}

ML models are increasingly deployed in settings with real world
interactions such as vehicles, but unfortunately, these models can fail in
systematic ways.
To prevent errors, ML engineering teams monitor and continuously improve these models.
We propose a new abstraction, \emph{model assertions}, that adapts the classical
use of program assertions as a way to monitor and improve ML models.
Model assertions are arbitrary functions over a model's input and output that
indicate when errors may be occurring, e.g., a function that triggers if an
object rapidly changes its class in a video.
We propose methods of using model assertions at all stages of ML system deployment,
including runtime monitoring, validating labels, and continuously improving ML models.
For runtime monitoring, we show that model assertions can find \emph{high confidence} errors,
where a model returns the wrong output with high confidence, which uncertainty-based
monitoring techniques would not detect.
For training, we propose two methods of using model assertions.
First, we propose a bandit-based active learning algorithm that can sample from
data flagged by assertions and show that it can reduce labeling costs by up to
40\%
over traditional uncertainty-based methods.
Second, we propose an API for generating ``consistency assertions'' (e.g., the
class change example)
and weak labels for inputs where the consistency assertions fail, and show
that these weak labels can improve relative model quality by
up to 46\%.
We evaluate model assertions on four real-world tasks with video, LIDAR, and ECG data.

\end{abstract}

]



\printAffiliationsAndNotice{\mlsysEqualContribution} 

\section{Introduction}

ML is increasingly deployed in \colora{complex 
contexts that require inference about the physical world}, from autonomous vehicles (AVs) to precision
medicine. However, ML models can \colora{misbehave in unexpected ways.} For example, AVs have accelerated toward highway
lane dividers~\cite{lee2018tesla} and can rapidly change their classification of objects
over time, causing erratic behavior~\cite{coldewey2018uber, ntsb2019vehicle}. As a
result, quality assurance (QA) of models, including continuous monitoring and
improvement, is of paramount concern.

Unfortunately, performing QA for complex, real-world ML applications is
challenging: ML models fail for diverse and reasons unknown before deployment.
Thus, existing solutions that focus on verifying training, including formal
verification~\cite{katz2017reluplex}, whitebox testing~\cite{pei2017deepxplore},
monitoring training metrics~\cite{renggli2019continuous}, and validating training
code~\cite{odena2018tensorfuzz}, 
\emph{only give guarantees on a test set and perturbations thereof}, so models
can still fail on the huge volumes of deployment data that are not part of the
test set (e.g., billions of images per day in an AV fleet).
Validating input schemas~\cite{polyzotis2019data, baylor2017tfx}
does not work for applications with unstructured inputs that \emph{lack
meaningful schemas}, e.g., images.
Solutions that check whether model performance remains consistent over
time~\cite{baylor2017tfx} only apply to deployments that have ground truth
labels, e.g., click-through rate prediction, but not to deployments that
\emph{lack labels}.

\begin{figure}[t!]
  \begin{subfigure}{.32\columnwidth}
    \includegraphics[width=0.99\columnwidth]{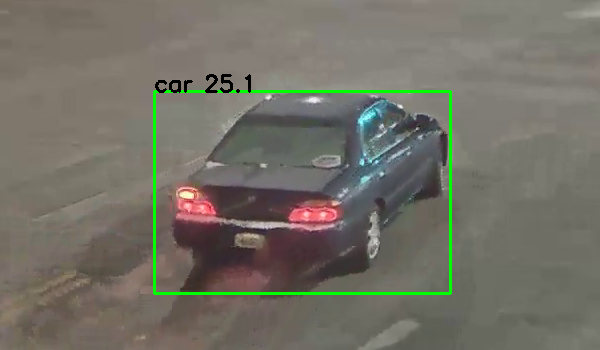}
    \caption{Frame 1, SSD}
  \end{subfigure}
  \begin{subfigure}{.32\columnwidth}
    \includegraphics[width=0.99\columnwidth]{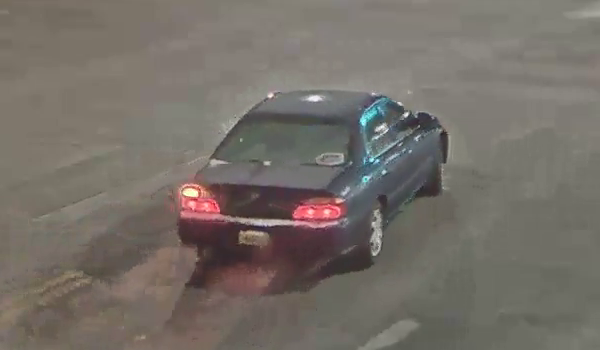}
    \caption{Frame 2, SSD}
  \end{subfigure}
  \begin{subfigure}{.32\columnwidth}
    \includegraphics[width=0.99\columnwidth]{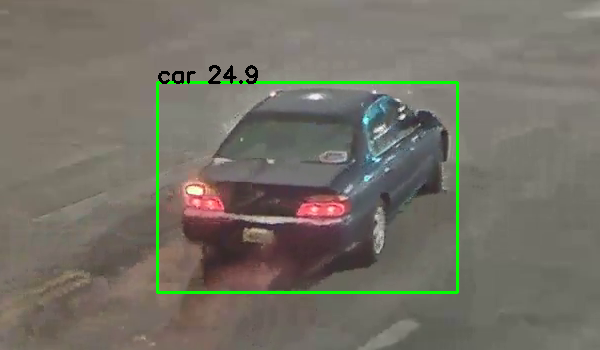}
    \caption{Frame 3, SSD}
  \end{subfigure}
  \begin{subfigure}{.32\columnwidth}
    \includegraphics[width=0.99\columnwidth]{figures/flicker/frame1.png}
    \caption{Frame 1, SSD\\\hspace{1in}}
  \end{subfigure}
  \begin{subfigure}{.32\columnwidth}
    \includegraphics[width=0.99\columnwidth]{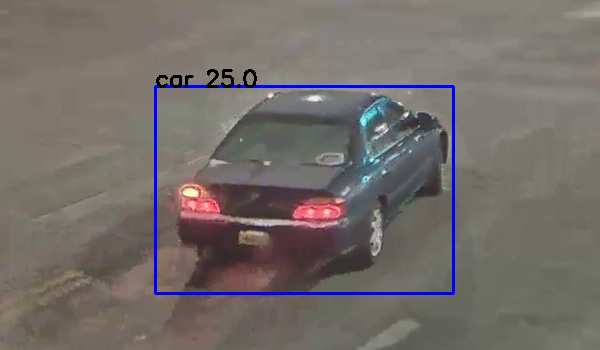}
    \caption{Frame 2, assertion corrected}
  \end{subfigure}
  \begin{subfigure}{.32\columnwidth}
    \includegraphics[width=0.99\columnwidth]{figures/flicker/frame3.png}
    \caption{Frame 3, SSD\\\hspace{1in}}
  \end{subfigure}
  \vspace{-0.5em}
  \caption{\textbf{Top row:} example of flickering in three consecutive frames
  of a video. The object detection method, SSD~\cite{liu2016ssd}, failed to
  identify the car in the second frame. \textbf{Bottom row:} example of
  correcting the output of a model. The car bounding box in the second frame can
  be inferred using nearby frames based on a consistency
  assertion.}
  \label{fig:flickering}
  \vspace{-0.5em}
\end{figure}

As a step towards more robust QA for complex ML applications, we have found
that ML developers can often specify \emph{systematic} errors made by
ML models: certain classes of errors are repetitive and can be
checked automatically, via code. For example, in developing a video analytics
engine, we noticed that object detection models can identify
boxes of cars that flicker rapidly in and out of the video (Figure~\ref{fig:flickering}),
indicating some of the detections are likely wrong.
Likewise, our contacts at an AV company reported that LIDAR and
camera models sometimes disagree.
While seemingly simple, similar errors were involved with a fatal AV crash \cite{ntsb2019vehicle}.
These systematic errors can arise for diverse reasons,
including domain shift between training and deployment data (e.g., still images
vs.~video), incomplete training data (e.g., no instances of snow-covered cars),
and noisy inputs.

To leverage the systematic nature of these errors, we propose \textit{model
assertions}, an abstraction to monitor and improve ML model quality.
Model assertions are inspired by program assertions~\cite{goldstine1947planning,
turing1949checking}, one of the most common ways to monitor software.
A model assertion is an arbitrary function over a model's input and output that
returns a Boolean (0 or 1) or continuous (floating point) severity score to indicate when faults may be
occurring.
For example, a model assertion that checks whether an object flickers in and
out of video could return a Boolean value over each frame or the number of objects that
flicker.
While assertions may not offer a complete specification of correctness, we have
found that assertions are easy to specify in many domains
(\S\ref{sec:model-assertions}).

We explore several ways to use model assertions, both at runtime and training
time.

First, we show that model assertions can be used for \textbf{runtime
monitoring}: they can be used to log unexpected behavior or automatically
trigger corrective actions, e.g., shutting down an autopilot. Furthermore, model
assertions can often find \emph{high confidence} errors, where the model has
high certainty in an erroneous output; these errors are problematic because
prior uncertainty-based monitoring would not flag these errors.
Additionally, and perhaps surprisingly, we have found that many groups are also
interested in validating human-generated labels, which can be done using model
assertions.

Second, we show that assertions can be used for \textbf{active learning}, in
which data is continuously collected to improve ML models.
Traditional active learning algorithms select data to label based on uncertainty,
with the intuition that ``harder'' data where the model is uncertain will be more
informative~\cite{settles2009active, coleman2020selection}.
Model assertions provide another natural way to find ``hard'' examples.
However, using assertions in active learning presents a challenge: how should
the active learning algorithm select between data when several assertions are
used? A data point can be flagged by multiple assertions or a
single assertion can flag multiple data points, in contrast to a single
uncertainty metric.
To address this challenge, we present a novel \emph{bandit-based active learning
algorithm (BAL)}.
Given a set of data that have been flagged by potentially multiple
model assertions, our bandit algorithm uses the assertions' severity scores as context
(i.e., features) and maximizes the marginal reduction in the number of assertions fired
(\S\ref{sec:alg}).
We show that our bandit algorithm can reduce labeling costs by up to 40\% over
traditional uncertainty-based methods.

Third, we show that assertions can be used for \textbf{weak supervision}~\cite{mintz2009distant, ratner2017weak}.
We propose an API for writing \emph{consistency assertions} about how attributes of
a model's output should relate that can also provide weak labels for training.
Consistency assertions specify that data should be consistent between attributes and
identifiers, e.g., a TV news host (identifier) should have consistent
gender (attribute), or that certain predictions should (or
should not) exist in temporally related outputs, e.g., cars in adjacent video frames
(Figure~\ref{fig:flickering}).
We demonstrate that this API can apply to a range of
domains, including medical classification and TV news analytics.
These weak labels can be used to improve relative model
quality by up to 46\% with no additional human labeling.

We implement model assertions in a Python library, \sn\footnote{OMG is a
recursive acronym for OMG Model Guardian.}, that
can be used with existing ML frameworks.
We evaluate assertions on four ML applications:
understanding TV news, AVs, video analytics, and classifying
medical readings.
We implement assertions for systematic errors reported by ML users in these
domains, including checking for consistency between sensors, domain knowledge
about object locations in videos, and medical knowledge about heart patterns.
Across these domains, we find that
model assertions we consider can be written with at most 60 lines of code and with
88-100\% precision, that these assertions often find high-confidence errors
(e.g., top 90th percentile by confidence), and
that our new algorithms for active learning and weak supervision via assertions
improve model quality over existing methods. 

In summary, we make the following contributions:
\begin{enumerate}[leftmargin=1em, topsep=-0.3em, itemsep=-0.5em]
  \item We introduce the abstraction of model assertions for monitoring and
  continuously improving ML models.
  
  \item We show that model assertions can find high confidence errors,
  which would not be flagged by uncertainty metrics.

  \item We propose a bandit algorithm to select data points for active learning via
  model assertions and show that it can reduce labeling costs by up to 40\%.

  \item We propose an API for consistency assertions that can automatically
  generate
  weak labels for data where the assertion fails, and show that weak supervision
  via these labels can improve relative model quality by up to 46\%.
\end{enumerate}


\section{Model Assertions}
\label{sec:model-assertions}

We describe the model assertion interface, examples of model assertions, how model assertions
can integrate into the ML development/deployment cycle, and its implementation
in \sn.

\subsection{Model Assertions Interface}
%

We formalize the model assertions interface. Model assertions
are arbitrary functions that can indicate when an error is likely to have
occurred.  They take as input a list of inputs and outputs from one or more ML
models. They return a \emph{severity score}, a continuous value that indicates the
severity of an error of a specific type. By convention, the \code{0} value
represents an abstention. Boolean values can be implemented in model assertions
by only returning \code{0} and \code{1}. The severity score does not need to
be calibrated, as our algorithms only use the relative ordering of scores.

As a concrete example, consider an AV with a LIDAR sensor and
camera and object detection models for each sensor. To check that these
models agree, a developer may write:
\begin{lstlisting}
def sensor_agreement(lidar_boxes, camera_boxes):
  failures = 0
  for lidar_box in lidar_boxes:
    if no_overlap(lidar_box, camera_boxes):
      failures += 1
  return failures
\end{lstlisting}
\vspace{-0.5em}
Notably, our library \sn can register arbitrary Python functions as model
assertions.

\subsection{Example Use Cases and Assertions}
\label{sec:use-cases}

In this section, we provide use cases for model assertions that arose in
discussions with industry and academic contacts, including AV companies and
academic labs. We show example of errors caught by the model assertions
described in this section in Appendix~\ref{sec:error-examples} and
describe how one might look for assertions in other domains in
Appendix~\ref{sec:ma-table}.

Our discussions revealed two key properties in real-world ML systems. First, ML
models are \emph{deployed on orders of magnitude more data than can reasonably be
labeled, so a labeled sample cannot capture all deployment conditions}. For
example, the fleet of Tesla vehicles will see over 100$\times$ more images in a
day than in the largest existing image dataset~\cite{sun2017revisiting}. Second,
complex ML deployments are developed by large teams, of which some developers
may not have the ability to manage all parts of the application. As a result, it
is critical to be able to do QA collaboratively to cover the application
end-to-end.

\minihead{Analyzing TV news}
We spoke to a research lab studying bias in media via automatic analysis.
This lab collected over 10 years of TV news (billions of frames) and
executed face detection every three seconds. These detections are subsequently
used to identify the faces, detect gender, and classify hair color using ML
models. Currently, the researchers have no method of identifying errors and
manually inspect data. However, they additionally compute scene cuts. Given that
most TV new hosts do not move much between scenes, we can assert that the
identity, gender, and hair color of faces that highly overlap within the same
scene are consistent (Figure~\ref{fig:tv-news}, Appendix). We further describe
how model assertions can be implemented via our consistency API for TV news in
\S\ref{sec:synth}.

\minihead{Autonomous vehicles (AVs)}
AVs are required to execute a variety of tasks, including detecting objects and
tracking lane markings. These tasks are accomplished with ML models from
different sensors, such as visual, LIDAR, or ultrasound
sensors~\cite{davies2018how}. For example, a vision model might be used to
detect objects in video and a point cloud model might be used to do 3D object
detection.

Our contacts at an AV company noticed that models from video and point
clouds can disagree. We implemented a model assertion that projects the 3D
boxes onto the 2D camera plane to check for consistency. If the assertion
triggers, then at least one of the sensors returned an incorrect answer.

\minihead{Video analytics}
Many modern, academic video analytics systems use an object detection
method~\cite{kang2017noscope, kang2018blazeit, hsieh2018focus,
jiang2018chameleon, xu2019vstore, canel2019scaling} trained on
MS-COCO~\cite{lin2014microsoft}, a corpus of still images. These still
image object detection methods are deployed on video for detecting objects.
None of these systems aim to detect errors, even though errors can affect
analytics results.

In developing such systems, we noticed that objects flicker in and out
of the video (Figure~\ref{fig:flickering}) and that vehicles overlap in
unrealistic ways (Figure~\ref{fig:multibox}, Appendix). We implemented
assertions to detect these.

\minihead{Medical classification}
Deep learning researchers have created deep networks that can outperform
cardiologists for classifying atrial fibrillation (AF, a form of heart
condition) from single-lead ECG data~\cite{rajpurkar2017cardiologist}. Our
researcher contacts mentioned that AF predictions from DNNs can rapidly oscillate.
The European Society of Cardiology guidelines for detecting AF require at least
30 seconds of signal before calling a detection~\cite{developed2010guidelines}.
Thus, predictions should not rapidly switch between two states. A developer could
specify this model assertion, which could be implemented to monitor ECG
classification deployments.

\begin{figure}
  \includegraphics[width=0.99\columnwidth]{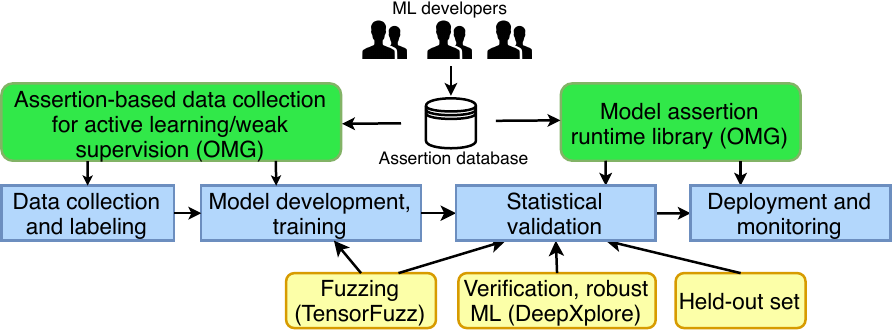}
  \vspace{-0.5em}
  \caption{A system diagram of how model assertions can integrate into the ML
  development/deployment pipeline. Users can collaboratively add to an assertion
  database. We also show how related work can be integrated into the pipeline.
  Notably, verification only gives guarantees on a test set and perturbations
  thereof, but not on arbitrary runtime data.}
  \vspace{-0.5em}
  \label{fig:omg-context}
\end{figure}

\subsection{Using Model Assertions for QA}

We describe how model assertions can be integrated with ML development and
deployment pipelines. Importantly, model assertions are complementary to a range
of other ML QA techniques, including verification, fuzzing, and statistical
techniques, as shown in Figure~\ref{fig:omg-context}.

First, model assertions can be used for monitoring and validating all parts of
the ML development/deployment pipeline. Namely, model assertions are agnostic to
the source of the output, whether they be ML models or human labelers. Perhaps
surprisingly, we have found several groups to also be interested in monitoring
human label quality. Thus, concretely, model assertions can be used to validate
human labels (data collection) or historical data (validation), and to monitor
deployments (e.g., to populate dashboards).

Second, model assertions can be used at training time to select which data points
to label in active learning. We describe BAL, our algorithm for data selection,
in \S\ref{sec:alg}.

Third, model assertions can be used to generate weak labels to further train ML
models without additional human labels. We describe how \sn accomplishes this
via consistency assertions in \S\ref{sec:synth}. Users can also register their
own weak supervision rules.

\subsection{Implementing Model Assertions in \sn}

We implement a prototype library for model assertions, \sn, that works with
existing Python ML training and deployment frameworks. We briefly describe \sn's
implementation.

\sn logs user-defined assertions as callbacks. The simplest way to
add an assertion is through \code{AddAssertion(func)}, where \code{func} is
a function of the inputs and outputs (see below). \sn also provides an API to
add consistency assertions as described in \S\ref{sec:synth}. Given this
database, \sn requires a callback after model execution that takes the model's
input and output as input. Given the model's input and output, \sn will execute
the assertions and record any errors. We assume the assertion signature is
similar to the following; this assertion signature is for the example in
Figure~\ref{fig:flickering}:
\begin{lstlisting}
def flickering(recent_frames: List[PixelBuf],
  recent_outputs: List[BoundingBox]) -> Float
\end{lstlisting}

For active learning, \sn will take a batch of data and return indices for which
data points to label. For weak supervision, \sn will take data and return weak
labels where valid. Users can specify weak labeling functions associated with
assertions to help with this.

In the following two sections, we describe two key methods that \sn uses to
improve model quality: BAL for active learning and consistency assertions for
weak supervision.

\section{Using Model Assertions for Active Learning with BAL}
\label{sec:alg}

We introduce an algorithm called BAL to select data for active learning via
model assertions. BAL assumes that a set of data points has been collected and a
subset will be labeled in bulk. We found that labeling services \cite{scale} and
our industrial contacts usually label data in bulk.

Given a set of data points that triggered model assertions, \sn must select
which points to label. There are two key challenges which make data
selection intractable in its full generality. First, we do not know the marginal
utility of selecting a data point to label without labeling the data point.
Second, even with labels, estimating the marginal gain of data points is
expensive to compute as training modern ML models is expensive.

To address these issues, we make simplifying assumptions. We describe the
statistical model we assume, the resource-unconstrained algorithm, our
simplifying assumptions, and BAL. We note that, while the resource-unconstrained
algorithm can produce statistical guarantees, BAL does not. We instead
empirically verify its performance in Section~\ref{sec:eval}.

\minihead{Data selection as multi-armed bandits}
We cast the data selection problem as a multi-armed bandit (MAB)
problem \cite{auer2002finite, berry1985bandit}. In MABs, a set of ``arms''
(i.e., individual data points) is provided and the user must select a set of
arms (i.e., points to label) to achieve the maximal expected utility (e.g.,
maximize validation accuracy, minimize number of assertions that fire). MABs
have been studied in a wide variety of settings \cite{radlinski2008learning,
lu2010contextual, bubeck2009pure}, but we assume that the arms have context
associated with them (i.e., severity scores from model assertions) and give
submodular rewards (defined below). The rewards are possibly time-varying.
We further assume there is an (unknown) smoothness parameter that determines the
similarity between arms of similar contexts (formally, the $\alpha$ in the
H{\"o}lder condition \cite{evans1998graduate}).
The following
presentation is inspired by \citet{chen2018contextual}.

Concretely, we assume the data will be labeled in $T$ rounds and denote the
rounds $t = 1, ..., T$. We refer to the set of $n$ data points as $N = \{1, ...,
n\}$. Each data point has a $d$ dimensional feature vector associated with it,
where $d$ is the number of model assertions. We refer to the feature vector as
$x_i^t$, where $i$ is the data point index and $t$ is the round index; from
here, we will refer to the data points as $x_i^t$. Each entry in a feature
vector is the severity score from a model assertion. The feature vectors can
change over time as the model predictions, and therefore assertions, change over
the course of training.

We assume there is a budget on the number of arms (i.e., data points to label),
$B^t$, at every round. The user must select a set of arms $S^t = \{ x_{s_1},
..., x_{s_{B^t}} \}$ such that $|S^t| \leq B^t$.  We assume that the
reward from the arms, $R(S^t)$, is submodular in $S^t$. Intuitively,
submodularity implies diminishing marginal returns: adding the 100th data point
will not improve the reward as much as adding the 10th data point. Formally, we
first define the marginal gain of adding an extra arm:
\begin{equation}
  \Delta R(\{m\}, A) = R(A \cup \{ m \}) - R(A).
  \label{eq:marginal-gain}
\end{equation}
where $A \subset N$ is a subset of arms and $m \in N$ is an additional arm such
that $m \not\in A$. The submodularity condition states that, for any $A \subset
C \subset N$ and $m \not\in C$
\begin{equation}
  \Delta R(\{m\}, A) \geq \Delta R(\{m\}, C).
\end{equation}

\minihead{Resource-unconstrained algorithm}
Assuming an infinite labeling and computational budget, we describe an algorithm
that selects data points to train on. Unfortunately, this algorithm is not
feasible as it requires labels for every point and training the ML model many
times.

If we assume that rewards for individual arms can be queried, then a recent
bandit algorithm, CC-MAB~\cite{chen2018contextual} can achieve a regret of $O(c
T^{\frac{2 \alpha d}{3 \alpha d}} \log(T))$ for $\alpha$ to be the smoothness
parameter. A regret bound is
the (asymptotic) difference with respect to an oracle algorithm. Briefly, CC-MAB
explores under-explored arms until it is confident that certain arms have
highest reward. Then, it greedily takes the highest reward arms. Full details
are given in~\cite{chen2018contextual} and summarized in
Algorithm~\ref{alg:ccmab}.

\begin{algorithm}[!t]
  \small
  \KwIn{$T$, $B^t$, $N$, $R$}
  \KwOut{choice of arms $S^t$ at rounds $1, ..., T$}
  \For{$t = 1, ..., T$}{
    \eIf{Underexplored arms}{
      Select arms $S^t$ from under-explored contexts at random
    }{
      Select arms $S^t$ by highest marginal gain (Eq.~\ref{eq:marginal-gain}): \\
      \For{$i = 1, ..., B^t$}{
        $S^t_i = \argmax_{j \in N \setminus S^t_{i-1}} \Delta R(\{j\},
        S^t_{i-1})$
      }
    }
  }
  \caption{A summary of the CC-MAB algorithm. CC-MAB first explores
  under-explored arms, then greedily selects arms with highest marginal gain.
  Full details are given in~\cite{chen2018contextual}.}
  \label{alg:ccmab}
\end{algorithm}

Unfortunately, CC-MAB requires access to an estimate of selecting a single arm.
Estimating the gain of a single arm requires a label and requires retraining and
reevaluating the model, which is computationally infeasible for
expensive-to-train ML models, especially modern deep networks.

\minihead{Resource-constrained algorithm}
We make simplifying assumptions and use these to modify CC-MAB for the
resource-constrained setting. Our simplifying assumptions are that 1) data
points with similar contexts (i.e., $x_i^t$) are interchangeable, 2) data points with higher
severity scores have higher expected marginal gain, and 3)
reducing the number of triggered assertions will increase accuracy.

Under these assumptions, we do not require an estimate of the marginal reward
for each arm. Instead, we can approximate the marginal gain from selecting arms
with similar contexts by the total number of these arms that were selected.
This has two benefits. First, we can train a model on a set of arms (i.e., data
points) in batches instead of adding single arms at a time. Second, we can
select data points of similar contexts at random, without having to compute its
marginal gain.

\begin{algorithm}[!t]
  \small
  \KwIn{$T$, $B^t$, $N$, $R$}
  \KwOut{choice of arms $S^t$ at rounds $1, ..., T$}
  \For{$t = 1, ..., T$}{
    \eIf{t = 0}{
      Select data points uniformly at random from the $d$ model assertions
    }{
      Compute the marginal reduction $r_m$ of the number of times model assertion $m =
      1, ..., d$ triggered from the previous round\;

      \If{all $r_m < 1\%$}{
        Fall back to baseline method\;
        continue\;
      }

      \For{$i = 1, ..., B^t$}{
        Select model assertion $m$ proportional to $r_m$\;
        Select $x_i$ that triggers $m$, sample proportional to severity score
        rank\;
        Add $x_i$ to $S^t$\;
      }
    }
  }
  \caption{BAL algorithm for data selection for continuous training. BAL samples
  from the assertions at random in the first round, then selects the assertions
  that result in highest marginal reduction in the number of assertions that
  fire in subsequent rounds. BAL will default to random sampling or uncertainty
  sampling if none of the assertions reduce.}
  \label{alg:resource-constrained}
\end{algorithm}

Leveraging these assumptions, we can simplify Algorithm~\ref{alg:ccmab} to
require less computation for training models and to not require labels for all
data points. Our algorithm is described in
Algorithm~\ref{alg:resource-constrained}. Briefly, we approximate the marginal
gain of selecting batches of arms and select arms proportional to the marginal
gain. We additionally allocate 25\% of the budget in each round to randomly
sample arms that triggered different model assertions, uniformly; this is
inspired by $\epsilon$-greedy algorithms \cite{tokic2011value}. This ensures
that no contexts (i.e., model assertions) are underexplored as training
progresses. Finally, in some cases (e.g., with noisy assertions), it may not be
possible to reduce the number of assertions that fire. In this case, BAL will
default to random sampling or uncertainty sampling, as specified by the user.

%

\section{Consistency~Assertions~and Weak Supervision}
\label{sec:synth}

Although developers can write arbitrary Python functions as model assertions
in \sn, we found that many assertions can be specified using an even simpler,
high-level abstraction that we called \emph{consistency assertions}.
This interface allows \sn to generate multiple Boolean model assertions from
a high-level description of the model's output, as well as automatic
\emph{correction rules} that propose new labels for data that fail the
assertion to enable weak supervision.

The key idea of consistency assertions is to specify which attributes
of a model's output are expected to match across many invocations to the model.
For example, consider a TV news application that tries to locate faces in
TV footage and then identify their name and gender (one of the real-world applications
we discussed in \S\ref{sec:use-cases}).
The ML developer may wish to assert that, within each video, each person should
consistently be assigned the same gender, and should appear on the screen at similar
positions on most nearby frames.
Consistency assertions let developers specify such requirements by providing
two functions:
\vspace{-0.5em}
\begin{itemize}[leftmargin=1em, itemsep=0pt]
  \item An \emph{identification function} that returns an identifier for each
  model output. For example, in our TV application, this could be the person's
  name as identified by the model.
  \item An \emph{attributes function} that returns a list of named attributes
  expected to be consistent for each identifier. In our example, this could
  return the gender attribute.
\end{itemize}

Given these two functions, \sn generates multiple Boolean assertions that
check whether the various attributes of outputs with a common identifier match.
In addition, it generates correction rules that can replace an inconsistent attribute
with a guess at that attribute's value based on other instances of the identifier
(we simply use the most common value).
By running the model and these generated assertions over unlabeled data, \sn can
thus \emph{automatically} generate weak labels for data points that do not satisfy
the consistency assertions.
\colora{Notably, \sn provides another way of producing labels for training that is
complementary to human-generated labels and other sources of weak labels. \sn is
especially suited for unstructured sources, e.g., video.}
We show in \S\ref{sec:eval} that these weak labels can automatically increase model quality.

\subsection{API Details}

The consistency assertions API supports ML applications that run over multiple
inputs $x_i$ and produce zero or more outputs $y_{i,j}$ for each input.
For example, each output could be an object detected in a video frame.
The user provides two functions over outputs $y_{i,j}$:
\vspace{-0.5em}
\begin{itemize}[leftmargin=1em, itemsep=0pt]
  \item $\mathrm{Id}(y_{i,j})$ returns an \emph{identifier}
  for the output $y_{i,j}$, which is simply an opaque value. 
  \item $\mathrm{Attrs}(y_{i,j})$ returns zero or more \emph{attributes}
  for the output $y_{i,j}$, which are key-value pairs. 
\end{itemize}

In addition to checking attributes, we found that many applications also expect
their identifiers to appear in a ``temporally consistent'' fashion, where objects do
not disappear and reappear too quickly. For example, one would expect cars
identified in the video to stay on the screen for multiple frames instead of
``flickering'' in and out in most cases.
To express this expectation, developers can provide a \emph{temporal consistency
threshold}, $T$, which specifies that each identifier should not appear or
disappear for intervals less than $T$ seconds.
For example, we might set $T$ to one second for TV footage that frequently
cuts across frames, or 30 seconds for an activity classification algorithm
that distinguishes between walking and biking.
The full API for adding a consistency assertion is therefore
$\mathrm{AddConsistencyAssertion}(\mathrm{Id}, ~ \mathrm{Attrs}, ~ T)$.

\minihead{Examples}
We briefly describe how one can use consistency assertions in several ML tasks
motivated in \S\ref{sec:use-cases}:

\emph{Face identification in TV footage:} This application uses multiple ML
models to detect faces in images, match them to identities, classify their
gender, and classifier their hair color. We can use the detected identity as our
$\mathrm{Id}$ function and gender/hair color as attributes.

\emph{Video analytics for traffic cameras:} This application aims to detect
vehicles in video street traffic, and suffers from problems such as flickering
or changing classifications for an object. The model's output is bounding boxes
with classes on each frame. Because we lack a globally unique identifier (e.g.,
license plate number) for each object, we can assign a new identifier for each
box that appears and assign the same identifier as it persists through the
video. We can treat the class as an attribute and set $T$ as well to detect
flickering.

\emph{Heart rhythm classification from ECGs:} In this application, domain
experts informed us that atrial fibrillation heart rhythms need to persist
for at least 30 seconds to be considered a problem. 
We used the detected class as our identifier and set $T$ to 30 seconds.

\subsection{Generating Assertions and Labels from the API}

Given the $\mathrm{Id}$, $\mathrm{Attrs}$, and $T$ values, \sn automatically
generates Boolean assertions to check for matching attributes and to check
that when an identifier appears in the data, it persists for at least $T$
seconds. These assertions are treated the same as user-provided ones in
the rest of the system.

\sn also automatically generates corrective rules that propose a new label
for outputs that do not match their identifier's other outputs on an attribute.
The default behavior is to propose the most common value of that attribute
(e.g., the class detected for an object on most frames), but users can also
provide a $\mathrm{WeakLabel}$ function to suggest an alternative based on
all of that object's outputs.

For temporal consistency constraints via $T$, \sn will assert by default that at most one
transition can occur within a $T$-second window; this can be overridden. For example, an identifier
appearing is valid, but an identifier appearing, disappearing, then appearing is
invalid. If
a violation occurs, \sn will propose to remove, modify, or add predictions. In the latter case,
\sn needs to know how to generate an expected output on an input where the
object was not identified (e.g., frames where the object flickered out in
Figure~\ref{fig:flickering}). \sn requires the user to provide a
$\mathrm{WeakLabel}$ function to cover this case, since it may require domain
specific logic, e.g., averaging the locations of the object on nearby video
frames.

\section{Evaluation}
\label{sec:eval}

\subsection{Experimental Setup}
We evaluated \sn and model assertions on four diverse ML workloads based on real
industrial and academic use-cases: analyzing TV news, video analytics,
autonomous vehicles, and medical classification. For each domain, we describe
the task, dataset, model, training procedure, and assertions. A summary is given
in Table~\ref{table:eval-tasks}.

\begin{table*}
\small
\begin{tabularx}{\linewidth}{llX}
  Task & Model & Assertions \\ \hline
  TV news & Custom & Consistency (\S\ref{sec:synth}, \texttt{news}) \\
  Object detection (video) & SSD~\cite{liu2016ssd}
      & Three vehicles should not highly overlap (\texttt{multibox}), identity
      consistency assertions (\texttt{flicker} and \texttt{appear}) \\
  Vehicle detection (AVs)  & Second~\cite{yan2018second}, SSD
      & Agreement of Point cloud and image detections (\texttt{agree}),
      \texttt{multibox} \\
  AF classification & ResNet~\cite{rajpurkar2017cardiologist}
      & Consistency assertion within a 30s time window (\texttt{ECG})
\end{tabularx}
\vspace{-0.5em}
\caption{A summary of tasks, models, and assertions used in our evaluation.}
\label{table:eval-tasks}
\vspace{-0.5em}
\end{table*}

\minihead{TV news}
Our contacts analyzing TV news provided us 50 hour-long segments that were known
to be problematic. They further provided pre-computed boxes of faces,
identities, and hair colors; this data was computed from a range of models and
sources, including hand-labeling, weak labels, and custom classifiers. We
implemented the consistency assertions described in \S\ref{sec:synth}. We were
unable to access the training code for this domain so were unable to perform
retraining experiments for this domain.

\minihead{Video analytics}
Many modern video analytics systems use object detection as a core
primitive~\cite{kang2017noscope, kang2018blazeit, hsieh2018focus,
jiang2018chameleon, xu2019vstore, canel2019scaling}, in which the task is to
localize and classify the objects in a frame of video. We focus on the object
detection portion of these systems. We used a ResNet-34 SSD~\cite{liu2016ssd}
(henceforth SSD) model pretrained on MS-COCO~\cite{lin2014microsoft}. We
deployed SSD for detecting vehicles in the \code{night-street} (i.e., \code{jackson}) video that is
commonly used \cite{kang2017noscope,
xu2019vstore, canel2019scaling, hsieh2018focus}. We used a separate day of video
for training and testing.

We deployed three model assertions: \code{multibox}, \code{flicker}, and
\code{appear}. The \code{multibox} assertion fires when three boxes highly
overlap (Figure~\ref{fig:multibox}, Appendix). The \code{flicker} and
\code{appear} assertions are implemented with our consistency API as described
in \S\ref{sec:synth}. 

\minihead{Autonomous vehicles}
We studied the problem of object detection for autonomous vehicles
using the NuScenes dataset~\cite{nuscenes2019}, which contains labeled LIDAR
point clouds and associated visual images. We split the data into separate
train, unlabeled, and test splits. We detected vehicles only. We use the
open-source Second model with PointPillars~\cite{yan2018second,
lang2019pointpillars} for LIDAR detections and SSD for visual detections. We
improve SSD via active learning and weak supervision in our experiments.

As NuScenes contains time-aligned point clouds and images, we deployed a custom
assertion for 2D and 3D boxes agreeing, and the \code{multibox} assertion. We
deployed a custom weak supervision rule that imputed boxes from the 3D
predictions. While other assertions could have been deployed (e.g.,
\code{flicker}), we found that the dataset was not sampled frequently enough
(at 2 Hz) for these assertions.

\minihead{Medical classification}
We studied the problem of classifying atrial fibrillation (AF) via ECG signals. We
used a convolutional network that was shown to outperform
cardiologists~\cite{rajpurkar2017cardiologist}. Unfortunately, the full dataset
used in~\cite{rajpurkar2017cardiologist} is not publicly available, so we used
the CINC17 dataset~\cite{cinc17}. CINC17 contains 8,528 data points that we
split into train, validation, unlabeled, and test splits.

We consulted with medical researchers and deployed an assertion that asserts
that the classification should not change between two classes in under a 30
second time period (i.e., the assertion fires when the classification changes
from $A \to B \to A$ within 30 seconds), as described in \S\ref{sec:synth}.

\subsection{Model~Assertions~can~be~Written~with High~Precision~and~Few~LOC}
We first asked whether model assertions could be written succinctly. To test this, we
implemented the model assertions described above and counted the lines of code
(LOC) necessary for each assertion. We count the LOC for the identity and
attribute functions for the consistency assertions (see
Table~\ref{table:eval-tasks} for a summary of assertions). We counted the LOC
with and without the shared helper functions (e.g., computing box overlap); we
double counted the helper functions when used between assertions. As we show in
Table~\ref{table:loc}, both consistency and domain-specific assertions can be
written in under 25 LOC excluding shared helper functions and under 60 LOC when
including helper functions. Thus, model assertions can be written with few LOC.

\begin{table}
\centering
\small
\begin{tabular}{lll}
Assertion         & LOC (no helpers) & LOC (inc. helpers) \\ \hline
\texttt{news}     & 7               & 39 \\
\texttt{ECG}      & 23              & 50 \\
\texttt{flicker}  & 18              & 60 \\
\texttt{appear}   & 18              & 35 \\ \hline
\texttt{multibox} & 14              & 28 \\
\texttt{agree}    & 11              & 28
\end{tabular}
\vspace{-0.5em}
\caption{Number of lines of code (LOC) for each assertion. Consistency
assertions are on the top and custom assertions are on the bottom. All
assertions could be written in under 60 LOC including helper functions, when
double counting between assertions. The assertion main body could be written in
under 25 LOC in all cases. The helper functions included utilities such as
computing the overlap between boxes.}
\vspace{-0.5em}
\label{table:loc}
\end{table}

\begin{table}
\centering
\small
\begin{tabular}{lcc}
                  & Precision & Precision \\
Assertion         & (identifier and output) & (model output only) \\ \hline
\texttt{news}     & 100\% & 100\% \\
\texttt{ECG}      & 100\% & 100\% \\
\texttt{flicker}  & 100\% & 96\% \\
\texttt{appear}   & 100\% & 88\% \\ \hline
\texttt{multibox} & N/A   & 100\% \\
\texttt{agree}    & N/A   & 98\% \\
\end{tabular}
\vspace{-0.5em}
\caption{Precision of our model assertions we deployed on 50 randomly selected
examples. The top are consistency assertions and the bottom are custom
assertions. We report both precision in the ML model outputs only and when
counting errors in the identification function and ML model outputs for
consistency assertions.
As shown, model assertions can be written with 88-100\% precision
across all domains when only counting errors in the model outputs.}
\vspace{-0.5em}
\label{table:precision}
\end{table}

We then asked whether model assertions could be written with high precision. To test
this, we randomly sampled 50 data points that triggered each assertion and
manually checked whether that data point had an incorrect output from the ML model.
The consistency assertions return clusters of data points (e.g.,
\code{appear}) and we report the precision for errors in both the identifier and
ML model outputs and only the ML model outputs. As we show in
Table~\ref{table:precision}, model assertions achieve at least 88\%
precision in all cases.

\subsection{Model~Assertions~can~Identify High-Confidence Errors}
We asked whether model assertions can identify high-confidence errors, or errors
where the model returns the wrong output with high confidence. High-confidence
errors are important to identify as confidence is used in downstream tasks, such
as analytics queries and actuation decisions~\cite{kang2017noscope,
kang2018blazeit, hsieh2018focus, chinchali2019network}. Furthermore, sampling
solutions that are based on confidence would be unable to identify these errors.

To determine whether model assertions could identify high confidence errors, we collected
the 10 data points with highest confidence error for each of the model
assertions deployed for video analytics. We then plotted the percentile of the
confidence among all the boxes for each error.

\begin{figure}
  \includegraphics[width=0.99\columnwidth]{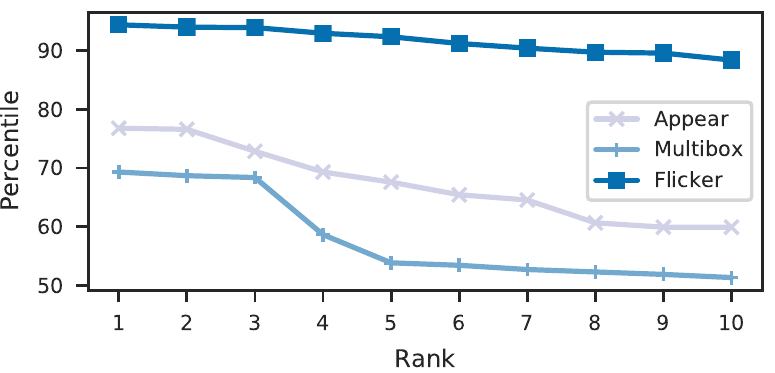}
  \vspace{-0.5em}
  \caption{Percentile of confidence of the top-10 ranked errors by confidence
  found by \sn for video analytics. The x-axis is the rank of the errors caught
  by model assertions, ordered by rank. The y-axis is the percentile of
  confidence among all the boxes. As shown, model assertions can find errors
  where the original model has high confidence (94th percentile), allowing them
  to complement existing confidence-based methods for data selection.}
  \vspace{-0.5em}
  \label{fig:percentiles-conf}
\end{figure}

As shown in Figure~\ref{fig:percentiles-conf}, model assertions can identify
errors within the top 94th percentile of boxes by confidence (the
\code{flicker} confidences were from the average of the surrounding boxes).
Importantly, uncertainty-based methods of monitoring would not catch these
errors.

We further show that model assertions can identify errors in human labels, which
effectively have a confidence of 1. These results are shown in
Appendix~\ref{sec:ma-human-labels}.

\subsection{Model~Assertions~can~Improve~Model~Quality via Active Learning}
We evaluated \sn's active learning capabilities and BAL using the three domains
for which we had access to the training code (visual analytics, ECG, AVs).

\minihead{Multiple model assertions}
We asked whether multiple model assertions could be used to improve model
quality via continuous data collection. We deployed three assertions over
\code{night-street} and two assertions for NuScenes. We used random sampling,
uncertainty sampling with ``least confident'' \cite{settles2009active}, \colora{uniform
sampling from data that triggered assertions,} and BAL for
the active learning strategies. We used the mAP metric for both datasets, which
is widely used for object detection~\cite{lin2014microsoft, he2017mask}. We
defer hyperparmeters to Appendix~\ref{sec:hyperparameters}.

\begin{figure}
  \begin{subfigure}{0.99\columnwidth}
    \includegraphics[width=0.99\columnwidth]{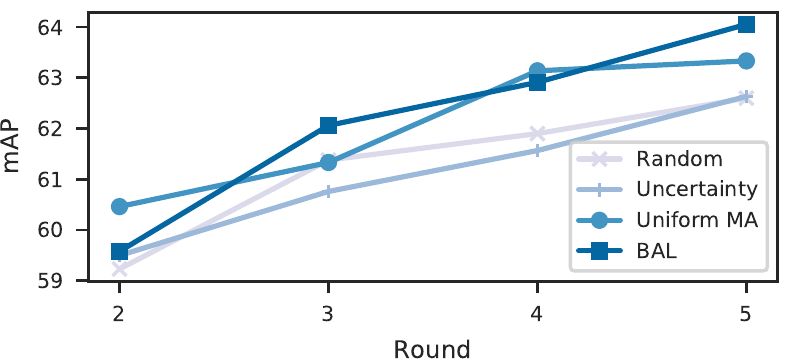}
    \vspace{-0.5em}
    \caption{Active learning for \texttt{night-street}.}
  \end{subfigure}
  \begin{subfigure}{0.99\columnwidth}
    \includegraphics[width=0.99\columnwidth]{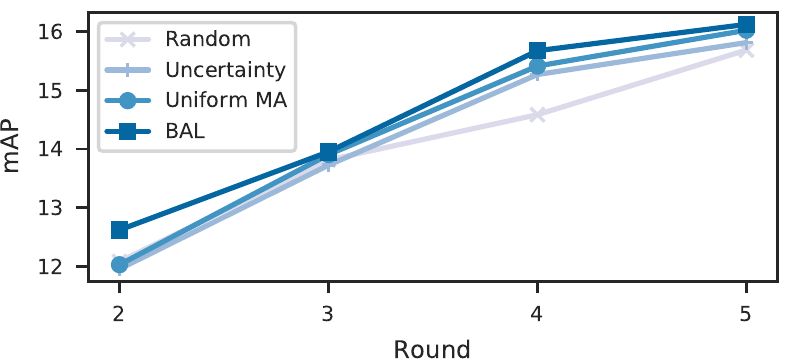}
    \vspace{-0.5em}
    \caption{Active learning for NuScenes.}
  \end{subfigure}
  \vspace{-0.5em}
  \caption{Performance of random sampling, uncertainty sampling, \colora{uniform
  sampling from model assertions,} and BAL for
  active learning. The round is the round of data collection (see
  \S\ref{sec:alg}). As shown in (a), BAL improves accuracy on unseen data and can
  achieve an accuracy target (62\% mAP) with 40\% fewer labels compared to
  random and uncertainty sampling
  for \texttt{night-street}. BAL also outperforms both baselines for the
  NuScenes dataset as shown in (b). We show figures with all rounds of active learning in
  Appendix~\ref{sec:full-al}.}
  \vspace{-0.5em}
  \label{fig:multi-active}
\end{figure}

As we show in Figure~\ref{fig:multi-active}, BAL outperforms both random
sampling and uncertainty sampling on both datasets after the first round, which
is required for calibration. \colora{BAL also outperforms uniform sampling from model
assertions by the last round.} For \code{night-street},
at a fixed accuracy threshold of 62\%, BAL uses 40\% fewer labels than random
and uncertainty sampling. By the fifth round, BAL outperforms both random
sampling and uncertainty sampling by 1.5\% mAP. While the absolute change in mAP
may seem small, doubling the model depth, which doubles the computational
budget, on MS-COCO achieves a 1.7\% improvement in mAP (ResNet-50 FPN vs.
ResNet-101 FPN)~\cite{Detectron2018}.

These results are expected, as prior work has shown that uncertainty sampling can
be unsuited for deep networks~\cite{sener2017active}.

\begin{figure}
  \includegraphics[width=0.99\columnwidth]{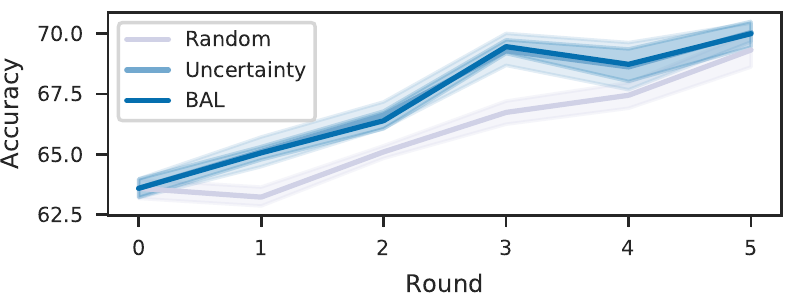}
  \vspace{-0.5em}
  \caption{Active learning results with a single assertion for the ECG dataset.
  As shown, with just a single assertion, model-assertion based active learning
  can match uncertainty sampling and outperform random sampling.}
  \vspace{-0.5em}
  \label{fig:single-active}
\end{figure}

\minihead{Single model assertion}
Due to the limited data quantities for the ECG dataset, we were unable to deploy
more than one assertion. Nonetheless, we further asked whether a single model
assertion could be used to improve model quality. We ran five rounds of data
labeling with 100 examples each round for ECG datasets. We ran the experiment 8
times and report averages. We show results in Figure~\ref{fig:single-active}. As
shown, data collection with a single model assertion generally matches or
outperforms both uncertainty and random sampling.

\subsection{Model~Assertions~can~Improve~Model~Quality via Weak Supervision}
We used our consistency assertions to evaluate the impact of weak supervision
using assertions for the domains we had weak labels for (video analytics, AVs,
and ECG).

\colora{
For \code{night-street}, we used 1,000 additional frames with 750 frames that
triggered \code{flicker} and 250 random frames with a learning rate of
$5\times10^{-6}$ for a total of 6 epochs. For the NuScenes dataset, we used the
same 350 scenes to bootstrap the LIDAR model as in the active learning
experiments. We trained with 175 scenes of weakly supervised data for one epoch
with a learning rate of $5 \times 10^{-5}$. For the ECG dataset, we used 1,000
weak labels and the same training procedure as in active learning.

}

\begin{table}
\centering
\small
\begin{tabular}{lll}
  Domain & Pretrained & Weakly supervised \\ \hline
  Video analytics (mAP) & 34.4 & 49.9 \\
  AVs (mAP)             & 10.6 & 14.1 \\
  ECG (\% accuracy)     & 70.7 & 72.1
\end{tabular}
\vspace{-0.5em}
\caption{Accuracy of the pretrained and weakly supervised models for video
analytics, AV and ECG domains. Weak supervision can improve accuracy with no
human-generated labels.}
\vspace{-0.5em}
\label{table:weak-supervision}
\end{table}

Table~\ref{table:weak-supervision} shows that model assertion-based weak
supervision can improve relative performance by 46.4\% for video analytics and
33\% for AVs. Similarly, the ECG classification can also improve with no
human-generated labels. These results show that model assertions can be useful
as a primitive for improving model quality with no additional data
labeling.

\section{Related Work}
\label{sec:related-work}

\minihead{ML QA}
A range of existing ML QA tools focus on validating
inputs via schemas or tracking
performance over time \cite{polyzotis2019data, baylor2017tfx}. However, these systems apply to
situations with meaningful schemas (e.g., tabular data) and ground-truth labels
at test time (e.g., predicting click-through rate). While model assertions could
also apply to these cases, they also cover situations that do not contain meaningful
schemas or labels at test time.

Other ML QA systems focus on training pipelines \cite{renggli2019continuous} or
validating numerical errors \cite{odena2018tensorfuzz}. These approaches are
important at finding pre-deployment bugs, but do not apply to test-time
scenarios; they are complementary to model assertions.

White-box testing systems, e.g., DeepXplore \cite{pei2017deepxplore},
test ML models by taking inputs and perturbing them. However, as
discussed, a validation set cannot cover all possibilities in the deployment
set. Furthermore, these systems do not give guarantees under model drift.

Since our initial workshop paper \cite{kang2018model}, several works have
extended model assertions \cite{arechiga2019better, henzinger2019outside}.

\minihead{Verified ML}
Verification has been applied to ML models in simple cases. For example,
Reluplex~\citep{katz2017reluplex} can verify that extremely small networks will
make correct control decisions given a fixed set of inputs and other work has
shown that similarly small networks can be verified against minimal
perturbations of a fixed set of input images~\citep{raghunathan2018certified}.
However, verification requires a specification, which may not be feasible to
implement, e.g., even humans may disagree on certain
predictions~\cite{kirillov2018panoptic}. Furthermore, the largest verified
networks we are aware of~\cite{katz2017reluplex, raghunathan2018certified,
wang2018formal, sun2019formal} are orders of magnitude smaller than the networks
we consider.

\minihead{Software Debugging}
Writing correct software and verifying software has a long
history, with many proposals from the research community. We hope that many such
practices are adopted in deploying machine learning models; we focus on
assertions in this work~\citep{goldstine1947planning, turing1949checking}.
Assertions have been shown to reduce the prevalence of bugs, when deployed
correctly~\citep{kudrjavets2006assessing, mahmood1984executable}. There are many
other such methods, such as formal verification~\citep{klein2009sel4,
leroy2009formal, keller1976formal}, conducting large-scale testing
(e.g., fuzzing)~\citep{takanen2008fuzzing, godefroid2012sage}, and symbolic
execution to trigger assertions~\citep{king1976symbolic, cadar2008klee}.
Probabilistic assertions have been used to verify simple distributional
properties of programs, such as differentially private programs should return an
expected mean~\cite{sampson2014expressing}. However, ML developers may not be
able to specify distributions and data may shift in deployment.

\minihead{Structured Prediction, Inductive Bias}
Several ML methods encode structure/inductive biases into training procedures or
models \citep{bakir2007predicting, haussler1988quantifying,
bakir2007predicting}. While promising, designing algorithms and models with specific
inductive biases can be challenging for non-experts. Additionally, these methods
generally do not contain runtime checks for aberrant behavior.

\minihead{Weak Supervision, Semi-supervised Learning}
Weak supervision leverages higher-level and/or noisier input
from human experts to improve model quality \citep{mintz2009distant,
ratner2017weak, jin2018unsupervised}. In semi-supervised learning, structural
assumptions over the data are used to leverage unlabeled
data~\citep{zhu2011semi}. However, to our knowledge, both of these methods do not contain
runtime checks and are not used in model-agnostic active
learning methods.


\section{Discussion}
\label{sec:discussion}

\colora{
While we believe model assertions are an important step towards
a practical solution for monitoring and continuously improving ML models, we
highlight three important limitations of model assertions, which may be fruitful
directions for future work.

First, certain model assertions may be difficult to express in our
current API. While arbitrary code can be expressed in \sn's API, certain
temporal assertions may be better expressed in a complex event processing
language~\cite{wu2006high}. We believe that domain-specific languages for model
assertions will be a fruitful area of future research.

Second, we have not thoroughly evaluated model assertions' performance in
real-time systems. Model assertions may add overhead to systems
where actuation has tight latency constraints, e.g., AVs. Nonetheless, model
assertions can be used over historical data for these systems. We are actively
collaborating with an AV company to explore these issues.

Third, certain issues in ML systems, such as bias in training sets, are out of
scope for model assertions. We hope that complementary systems, such as
TFX~\cite{baylor2017tfx}, can help improve quality in these cases.
}

\section{Conclusion}
\label{sec:conclusion}

In this work, we introduced model assertions, a model-agnostic technique that
allows domain experts to indicate errors in ML models. We showed that model
assertions can be used at runtime to detect high-confidence errors, which prior
methods would not detect. We proposed methods to use model assertions for active
learning and weak supervision to improve model quality. We implemented model
assertions in a novel library, \sn, and demonstrated that they can apply to a wide range of
real-world ML tasks, improving monitoring, active learning, and weak supervision
for ML models.

\section*{Acknowledgements}
{\small
This research was supported in part by affiliate members and other supporters of
the Stanford DAWN project---Ant Financial, Facebook, Google, Infosys, NEC, and
VMware---as well as Toyota Research Institute, Northrop Grumman, Cisco, SAP, and
the NSF under CAREER grant CNS-1651570 and Graduate Research Fellowship grant
DGE-1656518.
Any opinions, findings, and conclusions
or recommendations expressed in this material are those of the authors and do
not necessarily reflect the views of the National Science Foundation. Toyota
Research Institute (``TRI'') provided funds to assist the authors with their
research but this article solely reflects the opinions and conclusions of its
authors and not TRI or any other Toyota entity.

We further acknowledge Kayvon Fatahalian, James Hong, Dan Fu, Will Crichton,
Nikos Arechiga, and Sudeep Pillai for their productive discussions on ML
applications.
}

\bibliography{omg-sysml20}

\begin{thebibliography}{72}
\providecommand{\natexlab}[1]{#1}
\providecommand{\url}[1]{\texttt{#1}}
\expandafter\ifx\csname urlstyle\endcsname\relax
  \providecommand{\doi}[1]{doi: #1}\else
  \providecommand{\doi}{doi: \begingroup \urlstyle{rm}\Url}\fi

\bibitem[cin(2017)]{cinc17}
{AF} classification from a short single lead {ECG} recording: the
  physionet/computing in cardiology challenge 2017, 2017.
\newblock URL \url{https://physionet.org/challenge/2017/}.

\bibitem[sca(2019)]{scale}
Scale {API}: The {API} for training data, 2019.
\newblock URL \url{https://scale.ai/}.

\bibitem[Arechiga et~al.(2019)Arechiga, DeCastro, Kong, and
  Leung]{arechiga2019better}
Arechiga, N., DeCastro, J., Kong, S., and Leung, K.
\newblock Better {AI} through logical scaffolding.
\newblock \emph{arXiv preprint arXiv:1909.06965}, 2019.

\bibitem[Athalye et~al.(2018)Athalye, Engstrom, Ilyas, and
  Kwok]{athalye2018synthesizing}
Athalye, A., Engstrom, L., Ilyas, A., and Kwok, K.
\newblock Synthesizing robust adversarial examples.
\newblock \emph{ICML}, 2018.

\bibitem[Auer et~al.(2002)Auer, Cesa-Bianchi, and Fischer]{auer2002finite}
Auer, P., Cesa-Bianchi, N., and Fischer, P.
\newblock Finite-time analysis of the multiarmed bandit problem.
\newblock \emph{Machine learning}, 47\penalty0 (2-3):\penalty0 235--256, 2002.

\bibitem[BakIr et~al.(2007)BakIr, Hofmann, Sch{\"o}lkopf, Smola, Taskar, and
  Vishwanathan]{bakir2007predicting}
BakIr, G., Hofmann, T., Sch{\"o}lkopf, B., Smola, A.~J., Taskar, B., and
  Vishwanathan, S.
\newblock \emph{Predicting structured data}.
\newblock MIT press, 2007.

\bibitem[Baylor et~al.(2017)Baylor, Breck, Cheng, Fiedel, Foo, Haque, Haykal,
  Ispir, Jain, Koc, et~al.]{baylor2017tfx}
Baylor, D., Breck, E., Cheng, H.-T., Fiedel, N., Foo, C.~Y., Haque, Z., Haykal,
  S., Ispir, M., Jain, V., Koc, L., et~al.
\newblock Tfx: A tensorflow-based production-scale machine learning platform.
\newblock In \emph{SIGKDD}. ACM, 2017.

\bibitem[Berry \& Fristedt(1985)Berry and Fristedt]{berry1985bandit}
Berry, D.~A. and Fristedt, B.
\newblock Bandit problems: sequential allocation of experiments (monographs on
  statistics and applied probability).
\newblock \emph{London: Chapman and Hall}, 5:\penalty0 71--87, 1985.

\bibitem[Bubeck et~al.(2009)Bubeck, Munos, and Stoltz]{bubeck2009pure}
Bubeck, S., Munos, R., and Stoltz, G.
\newblock Pure exploration in multi-armed bandits problems.
\newblock In \emph{International conference on Algorithmic learning theory},
  pp.\  23--37. Springer, 2009.

\bibitem[Cadar et~al.(2008)Cadar, Dunbar, Engler, et~al.]{cadar2008klee}
Cadar, C., Dunbar, D., Engler, D.~R., et~al.
\newblock Klee: Unassisted and automatic generation of high-coverage tests for
  complex systems programs.
\newblock In \emph{OSDI}, volume~8, pp.\  209--224, 2008.

\bibitem[Caesar et~al.(2019)Caesar, Bankiti, Lang, Vora, Liong, Xu, Krishnan,
  Pan, Baldan, and Beijbom]{nuscenes2019}
Caesar, H., Bankiti, V., Lang, A.~H., Vora, S., Liong, V.~E., Xu, Q., Krishnan,
  A., Pan, Y., Baldan, G., and Beijbom, O.
\newblock nuscenes: A multimodal dataset for autonomous driving.
\newblock \emph{arXiv preprint arXiv:1903.11027}, 2019.

\bibitem[Canel et~al.(2019)Canel, Kim, Zhou, Li, Lim, Andersen, Kaminsky, and
  Dulloor]{canel2019scaling}
Canel, C., Kim, T., Zhou, G., Li, C., Lim, H., Andersen, D., Kaminsky, M., and
  Dulloor, S.
\newblock Scaling video analytics on constrained edge nodes.
\newblock \emph{SysML}, 2019.

\bibitem[Chen et~al.(2018)Chen, Xu, and Lu]{chen2018contextual}
Chen, L., Xu, J., and Lu, Z.
\newblock Contextual combinatorial multi-armed bandits with volatile arms and
  submodular reward.
\newblock In \emph{Advances in Neural Information Processing Systems}, pp.\
  3247--3256, 2018.

\bibitem[Chinchali et~al.(2019)Chinchali, Sharma, Harrison, Elhafsi, Kang,
  Pergament, Cidon, Katti, and Pavone]{chinchali2019network}
Chinchali, S., Sharma, A., Harrison, J., Elhafsi, A., Kang, D., Pergament, E.,
  Cidon, E., Katti, S., and Pavone, M.
\newblock Network offloading policies for cloud robotics: a learning-based
  approach.
\newblock \emph{arXiv preprint arXiv:1902.05703}, 2019.

\bibitem[Coldewey(2018)]{coldewey2018uber}
Coldewey, D.
\newblock Uber in fatal crash detected pedestrian but had emergency braking
  disabled, 2018.
\newblock URL
  \url{https://techcrunch.com/2018/05/24/uber-in-fatal-crash-detected-pedestrian-but-had-emergency-braking-disabled/}.

\bibitem[Coleman et~al.(2020)Coleman, Yeh, Mussmann, Mirzasoleiman, Bailis,
  Liang, Leskovec, and Zaharia]{coleman2020selection}
Coleman, C., Yeh, C., Mussmann, S., Mirzasoleiman, B., Bailis, P., Liang, P.,
  Leskovec, J., and Zaharia, M.
\newblock Selection via proxy: Efficient data selection for deep learning.
\newblock In \emph{International Conference on Learning Representations}, 2020.
\newblock URL \url{https://openreview.net/forum?id=HJg2b0VYDr}.

\bibitem[Davies(2018)]{davies2018how}
Davies, A.
\newblock How do self-driving cars see? (and how do they see me?), 2018.
\newblock URL
  \url{https://www.wired.com/story/the-know-it-alls-how-do-self-driving-cars-see/}.

\bibitem[EHRA(2010)]{developed2010guidelines}
EHRA.
\newblock Guidelines for the management of atrial fibrillation: the task force
  for the management of atrial fibrillation of the {E}uropean society of
  cardiology ({ESC}).
\newblock \emph{European heart journal}, 31\penalty0 (19):\penalty0 2369--2429,
  2010.

\bibitem[Evans(1998)]{evans1998graduate}
Evans, L.~C.
\newblock Graduate studies in mathematics.
\newblock In \emph{Partial differential equations}. Am. Math. Soc., 1998.

\bibitem[Girshick et~al.(2018)Girshick, Radosavovic, Gkioxari, Doll\'{a}r, and
  He]{Detectron2018}
Girshick, R., Radosavovic, I., Gkioxari, G., Doll\'{a}r, P., and He, K.
\newblock Detectron.
\newblock \url{https://github.com/facebookresearch/detectron}, 2018.

\bibitem[Godefroid et~al.(2012)Godefroid, Levin, and Molnar]{godefroid2012sage}
Godefroid, P., Levin, M.~Y., and Molnar, D.
\newblock Sage: whitebox fuzzing for security testing.
\newblock \emph{Queue}, 10\penalty0 (1):\penalty0 20, 2012.

\bibitem[Goldstine et~al.(1947)Goldstine, Von~Neumann, and
  Von~Neumann]{goldstine1947planning}
Goldstine, H.~H., Von~Neumann, J., and Von~Neumann, J.
\newblock Planning and coding of problems for an electronic computing
  instrument.
\newblock 1947.

\bibitem[Goodfellow et~al.(2015)Goodfellow, Shlens, and
  Szegedy]{goodfellow2015explaining}
Goodfellow, I.~J., Shlens, J., and Szegedy, C.
\newblock Explaining and harnessing adversarial examples.
\newblock \emph{ICLR}, 2015.

\bibitem[Haussler(1988)]{haussler1988quantifying}
Haussler, D.
\newblock Quantifying inductive bias: {AI} learning algorithms and valiant's
  learning framework.
\newblock \emph{Artificial intelligence}, 36\penalty0 (2):\penalty0 177--221,
  1988.

\bibitem[He et~al.(2017)He, Gkioxari, Doll{\'a}r, and Girshick]{he2017mask}
He, K., Gkioxari, G., Doll{\'a}r, P., and Girshick, R.
\newblock Mask r-cnn.
\newblock In \emph{Computer Vision (ICCV), 2017 IEEE International Conference
  on}, pp.\  2980--2988. IEEE, 2017.

\bibitem[Henzinger et~al.(2019)Henzinger, Lukina, and
  Schilling]{henzinger2019outside}
Henzinger, T.~A., Lukina, A., and Schilling, C.
\newblock Outside the box: Abstraction-based monitoring of neural networks.
\newblock \emph{arXiv preprint arXiv:1911.09032}, 2019.

\bibitem[Hirth et~al.(2013)Hirth, Ho{\ss}feld, and
  Tran-Gia]{hirth2013analyzing}
Hirth, M., Ho{\ss}feld, T., and Tran-Gia, P.
\newblock Analyzing costs and accuracy of validation mechanisms for
  crowdsourcing platforms.
\newblock \emph{Mathematical and Computer Modelling}, 57\penalty0
  (11-12):\penalty0 2918--2932, 2013.

\bibitem[Hsieh et~al.(2018)Hsieh, Ananthanarayanan, Bodik, Venkataraman, Bahl,
  Philipose, Gibbons, and Mutlu]{hsieh2018focus}
Hsieh, K., Ananthanarayanan, G., Bodik, P., Venkataraman, S., Bahl, P.,
  Philipose, M., Gibbons, P.~B., and Mutlu, O.
\newblock Focus: Querying large video datasets with low latency and low cost.
\newblock In \emph{OSDI}, pp.\  269--286, 2018.

\bibitem[Jiang et~al.(2018)Jiang, Ananthanarayanan, Bodik, Sen, and
  Stoica]{jiang2018chameleon}
Jiang, J., Ananthanarayanan, G., Bodik, P., Sen, S., and Stoica, I.
\newblock Chameleon: scalable adaptation of video analytics.
\newblock In \emph{Proceedings of the 2018 Conference of the ACM Special
  Interest Group on Data Communication}, pp.\  253--266. ACM, 2018.

\bibitem[Jin et~al.(2018)Jin, RoyChowdhury, Jiang, Singh, Prasad, Chakraborty,
  and Learned-Miller]{jin2018unsupervised}
Jin, S., RoyChowdhury, A., Jiang, H., Singh, A., Prasad, A., Chakraborty, D.,
  and Learned-Miller, E.
\newblock Unsupervised hard example mining from videos for improved object
  detection.
\newblock In \emph{Proceedings of the European Conference on Computer Vision
  (ECCV)}, pp.\  307--324, 2018.

\bibitem[Kang et~al.(2017)Kang, Emmons, Abuzaid, Bailis, and
  Zaharia]{kang2017noscope}
Kang, D., Emmons, J., Abuzaid, F., Bailis, P., and Zaharia, M.
\newblock Noscope: optimizing neural network queries over video at scale.
\newblock \emph{Proceedings of the VLDB Endowment}, 10\penalty0 (11):\penalty0
  1586--1597, 2017.

\bibitem[Kang et~al.(2018)Kang, Raghavan, Bailis, and Zaharia]{kang2018model}
Kang, D., Raghavan, D., Bailis, P., and Zaharia, M.
\newblock Model assertions for debugging machine learning.
\newblock In \emph{NeurIPS MLSys Workshop}, 2018.

\bibitem[Kang et~al.(2019)Kang, Bailis, and Zaharia]{kang2018blazeit}
Kang, D., Bailis, P., and Zaharia, M.
\newblock Blazeit: Fast exploratory video queries using neural networks.
\newblock \emph{PVLDB}, 2019.

\bibitem[Katz et~al.(2017)Katz, Barrett, Dill, Julian, and
  Kochenderfer]{katz2017reluplex}
Katz, G., Barrett, C., Dill, D.~L., Julian, K., and Kochenderfer, M.~J.
\newblock Reluplex: An efficient {SMT} solver for verifying deep neural
  networks.
\newblock In \emph{International Conference on Computer Aided Verification},
  pp.\  97--117. Springer, 2017.

\bibitem[Keller(1976)]{keller1976formal}
Keller, R.~M.
\newblock Formal verification of parallel programs.
\newblock \emph{Communications of the ACM}, 19\penalty0 (7):\penalty0 371--384,
  1976.

\bibitem[King(1976)]{king1976symbolic}
King, J.~C.
\newblock Symbolic execution and program testing.
\newblock \emph{Communications of the ACM}, 19\penalty0 (7):\penalty0 385--394,
  1976.

\bibitem[Kirillov et~al.(2018)Kirillov, He, Girshick, Rother, and
  Doll{\'a}r]{kirillov2018panoptic}
Kirillov, A., He, K., Girshick, R., Rother, C., and Doll{\'a}r, P.
\newblock Panoptic segmentation.
\newblock \emph{arXiv preprint arXiv:1801.00868}, 2018.

\bibitem[Klein et~al.(2009)Klein, Elphinstone, Heiser, Andronick, Cock, Derrin,
  Elkaduwe, Engelhardt, Kolanski, Norrish, et~al.]{klein2009sel4}
Klein, G., Elphinstone, K., Heiser, G., Andronick, J., Cock, D., Derrin, P.,
  Elkaduwe, D., Engelhardt, K., Kolanski, R., Norrish, M., et~al.
\newblock sel4: Formal verification of an {OS} kernel.
\newblock In \emph{Proceedings of the ACM SIGOPS 22nd symposium on Operating
  systems principles}, pp.\  207--220. ACM, 2009.

\bibitem[Kudrjavets et~al.(2006)Kudrjavets, Nagappan, and
  Ball]{kudrjavets2006assessing}
Kudrjavets, G., Nagappan, N., and Ball, T.
\newblock Assessing the relationship between software assertions and faults: An
  empirical investigation.
\newblock In \emph{Software Reliability Engineering, 2006. ISSRE'06. 17th
  International Symposium on}, pp.\  204--212. IEEE, 2006.

\bibitem[Lang et~al.(2019)Lang, Vora, Caesar, Zhou, Yang, and
  Beijbom]{lang2019pointpillars}
Lang, A.~H., Vora, S., Caesar, H., Zhou, L., Yang, J., and Beijbom, O.
\newblock Pointpillars: Fast encoders for object detection from point clouds.
\newblock In \emph{Proceedings of the IEEE Conference on Computer Vision and
  Pattern Recognition}, pp.\  12697--12705, 2019.

\bibitem[Lee(2018)]{lee2018tesla}
Lee, T.
\newblock Tesla says autopilot was active during fatal crash in mountain view.
\newblock
  \url{https://arstechnica.com/cars/2018/03/tesla-says-autopilot-was-active-during-fatal-crash-in-mountain-view/},
  2018.

\bibitem[Leroy(2009)]{leroy2009formal}
Leroy, X.
\newblock Formal verification of a realistic compiler.
\newblock \emph{Communications of the ACM}, 52\penalty0 (7):\penalty0 107--115,
  2009.

\bibitem[Lin et~al.(2014)Lin, Maire, Belongie, Hays, Perona, Ramanan,
  Doll{\'a}r, and Zitnick]{lin2014microsoft}
Lin, T.-Y., Maire, M., Belongie, S., Hays, J., Perona, P., Ramanan, D.,
  Doll{\'a}r, P., and Zitnick, C.~L.
\newblock Microsoft {COCO}: Common objects in context.
\newblock In \emph{European conference on computer vision}, pp.\  740--755.
  Springer, 2014.

\bibitem[Liu et~al.(2016)Liu, Anguelov, Erhan, Szegedy, Reed, Fu, and
  Berg]{liu2016ssd}
Liu, W., Anguelov, D., Erhan, D., Szegedy, C., Reed, S., Fu, C.-Y., and Berg,
  A.~C.
\newblock {SSD}: Single shot multibox detector.
\newblock In \emph{European conference on computer vision}, pp.\  21--37.
  Springer, 2016.

\bibitem[Lu et~al.(2010)Lu, P{\'a}l, and P{\'a}l]{lu2010contextual}
Lu, T., P{\'a}l, D., and P{\'a}l, M.
\newblock Contextual multi-armed bandits.
\newblock In \emph{Proceedings of the Thirteenth international conference on
  Artificial Intelligence and Statistics}, pp.\  485--492, 2010.

\bibitem[Mahmood et~al.(1984)Mahmood, Andrews, and
  McCluskey]{mahmood1984executable}
Mahmood, A., Andrews, D.~M., and McCluskey, E.~J.
\newblock Executable assertions and flight software.
\newblock 1984.

\bibitem[Mintz et~al.(2009)Mintz, Bills, Snow, and Jurafsky]{mintz2009distant}
Mintz, M., Bills, S., Snow, R., and Jurafsky, D.
\newblock Distant supervision for relation extraction without labeled data.
\newblock In \emph{Proceedings of the Joint Conference of the 47th Annual
  Meeting of the ACL and the 4th International Joint Conference on Natural
  Language Processing of the AFNLP: Volume 2-Volume 2}, pp.\  1003--1011.
  Association for Computational Linguistics, 2009.

\bibitem[{NTSB}(2019)]{ntsb2019vehicle}
{NTSB}.
\newblock Vehicle automation report, {HWY18MH010}, 2019.
\newblock URL \url{https://dms.ntsb.gov/public/62500-62999/62978/629713.pdf}.

\bibitem[Odena \& Goodfellow(2018)Odena and Goodfellow]{odena2018tensorfuzz}
Odena, A. and Goodfellow, I.
\newblock Tensorfuzz: Debugging neural networks with coverage-guided fuzzing.
\newblock \emph{arXiv preprint arXiv:1807.10875}, 2018.

\bibitem[Pei et~al.(2017)Pei, Cao, Yang, and Jana]{pei2017deepxplore}
Pei, K., Cao, Y., Yang, J., and Jana, S.
\newblock Deepxplore: Automated whitebox testing of deep learning systems.
\newblock In \emph{Proceedings of the 26th Symposium on Operating Systems
  Principles}, pp.\  1--18. ACM, 2017.

\bibitem[Polyzotis et~al.(2019)Polyzotis, Zinkevich, Roy, Breck, and
  Whang]{polyzotis2019data}
Polyzotis, N., Zinkevich, M., Roy, S., Breck, E., and Whang, S.
\newblock Data validation for machine learning.
\newblock \emph{SysML}, 2019.

\bibitem[Radlinski et~al.(2008)Radlinski, Kleinberg, and
  Joachims]{radlinski2008learning}
Radlinski, F., Kleinberg, R., and Joachims, T.
\newblock Learning diverse rankings with multi-armed bandits.
\newblock In \emph{Proceedings of the 25th international conference on Machine
  learning}, pp.\  784--791. ACM, 2008.

\bibitem[Raghunathan et~al.(2018)Raghunathan, Steinhardt, and
  Liang]{raghunathan2018certified}
Raghunathan, A., Steinhardt, J., and Liang, P.
\newblock Certified defenses against adversarial examples.
\newblock \emph{arXiv preprint arXiv:1801.09344}, 2018.

\bibitem[Rajpurkar et~al.(2019)Rajpurkar, Hannun, Haghpanahi, Bourn, and
  Ng]{rajpurkar2017cardiologist}
Rajpurkar, P., Hannun, A.~Y., Haghpanahi, M., Bourn, C., and Ng, A.~Y.
\newblock Cardiologist-level arrhythmia detection with convolutional neural
  networks.
\newblock \emph{Nature Medicine}, 2019.

\bibitem[Ratner et~al.(2017)Ratner, Bach, Varma, and R{\'e}]{ratner2017weak}
Ratner, A., Bach, S., Varma, P., and R{\'e}, C.
\newblock Weak supervision: The new programming paradigm for machine learning,
  2017.
\newblock URL \url{https://dawn.cs.stanford.edu/2017/07/16/weak-supervision/}.

\bibitem[Renggli et~al.(2019)Renggli, Karlaš, Ding, Liu, Schawinski, Wu, and
  Zhang]{renggli2019continuous}
Renggli, C., Karlaš, B., Ding, B., Liu, F., Schawinski, K., Wu, W., and Zhang,
  C.
\newblock Continuous integration of machine learning models with ease.ml/ci:
  Towards a rigorous yet practical treatment.
\newblock \emph{SysML}, 2019.

\bibitem[Sampson et~al.(2014)Sampson, Panchekha, Mytkowicz, McKinley, Grossman,
  and Ceze]{sampson2014expressing}
Sampson, A., Panchekha, P., Mytkowicz, T., McKinley, K.~S., Grossman, D., and
  Ceze, L.
\newblock Expressing and verifying probabilistic assertions.
\newblock \emph{ACM SIGPLAN Notices}, 49\penalty0 (6):\penalty0 112--122, 2014.

\bibitem[Sener \& Savarese(2017)Sener and Savarese]{sener2017active}
Sener, O. and Savarese, S.
\newblock Active learning for convolutional neural networks: A core-set
  approach.
\newblock \emph{arXiv preprint arXiv:1708.00489}, 2017.

\bibitem[Settles(2009)]{settles2009active}
Settles, B.
\newblock Active learning literature survey.
\newblock Technical report, University of Wisconsin-Madison Department of
  Computer Sciences, 2009.

\bibitem[Sun et~al.(2017)Sun, Shrivastava, Singh, and Gupta]{sun2017revisiting}
Sun, C., Shrivastava, A., Singh, S., and Gupta, A.
\newblock Revisiting unreasonable effectiveness of data in deep learning era.
\newblock In \emph{Proceedings of the IEEE international conference on computer
  vision}, pp.\  843--852, 2017.

\bibitem[Sun et~al.(2019)Sun, Khedr, and Shoukry]{sun2019formal}
Sun, X., Khedr, H., and Shoukry, Y.
\newblock Formal verification of neural network controlled autonomous systems.
\newblock In \emph{Proceedings of the 22nd ACM International Conference on
  Hybrid Systems: Computation and Control}, pp.\  147--156. ACM, 2019.

\bibitem[Takanen et~al.(2008)Takanen, Demott, and Miller]{takanen2008fuzzing}
Takanen, A., Demott, J.~D., and Miller, C.
\newblock \emph{Fuzzing for software security testing and quality assurance}.
\newblock Artech House, 2008.

\bibitem[Taylor \& Nitschke(2017)Taylor and Nitschke]{taylor2017improving}
Taylor, L. and Nitschke, G.
\newblock Improving deep learning using generic data augmentation.
\newblock \emph{arXiv preprint arXiv:1708.06020}, 2017.

\bibitem[Tokic \& Palm(2011)Tokic and Palm]{tokic2011value}
Tokic, M. and Palm, G.
\newblock Value-difference based exploration: adaptive control between
  epsilon-greedy and softmax.
\newblock In \emph{Annual Conference on Artificial Intelligence}, pp.\
  335--346. Springer, 2011.

\bibitem[Tran-Thanh et~al.(2013)Tran-Thanh, Venanzi, Rogers, and
  Jennings]{tran2013efficient}
Tran-Thanh, L., Venanzi, M., Rogers, A., and Jennings, N.~R.
\newblock Efficient budget allocation with accuracy guarantees for
  crowdsourcing classification tasks.
\newblock In \emph{Proceedings of the 2013 international conference on
  Autonomous agents and multi-agent systems}, pp.\  901--908. International
  Foundation for Autonomous Agents and Multiagent Systems, 2013.

\bibitem[Turing(1949)]{turing1949checking}
Turing, A.
\newblock Checking a large routine.
\newblock In \emph{Report on a Conference on High Speed Automatic Calculating
  machines}, pp.\  67--69. Cambridge University Mathematics Lab, 1949.

\bibitem[Wang \& Perez(2017)Wang and Perez]{wang2017effectiveness}
Wang, J. and Perez, L.
\newblock The effectiveness of data augmentation in image classification using
  deep learning.
\newblock \emph{Convolutional Neural Networks Vis. Recognit}, 2017.

\bibitem[Wang et~al.(2018)Wang, Pei, Whitehouse, Yang, and
  Jana]{wang2018formal}
Wang, S., Pei, K., Whitehouse, J., Yang, J., and Jana, S.
\newblock Formal security analysis of neural networks using symbolic intervals.
\newblock In \emph{USENIX Security Symposium}, pp.\  1599--1614, 2018.

\bibitem[Wu et~al.(2006)Wu, Diao, and Rizvi]{wu2006high}
Wu, E., Diao, Y., and Rizvi, S.
\newblock High-performance complex event processing over streams.
\newblock In \emph{Proceedings of the 2006 ACM SIGMOD international conference
  on Management of data}, pp.\  407--418. ACM, 2006.

\bibitem[Xu et~al.(2019)Xu, Botelho, and Lin]{xu2019vstore}
Xu, T., Botelho, L.~M., and Lin, F.~X.
\newblock Vstore: A data store for analytics on large videos.
\newblock In \emph{Proceedings of the Fourteenth EuroSys Conference 2019}, pp.\
  ~16. ACM, 2019.

\bibitem[Yan et~al.(2018)Yan, Mao, and Li]{yan2018second}
Yan, Y., Mao, Y., and Li, B.
\newblock Second: Sparsely embedded convolutional detection.
\newblock \emph{Sensors}, 18\penalty0 (10):\penalty0 3337, 2018.

\bibitem[Zhu(2011)]{zhu2011semi}
Zhu, X.
\newblock Semi-supervised learning.
\newblock In \emph{Encyclopedia of machine learning}, pp.\  892--897. Springer,
  2011.

\end{thebibliography}
\bibliographystyle{mlsys2020}

\clearpage


\appendix

\section{Examples of Errors Caught by Model Assertions}
\label{sec:error-examples}

\begin{figure}
  \begin{subfigure}{0.49\columnwidth}
    \includegraphics[width=0.98\columnwidth]{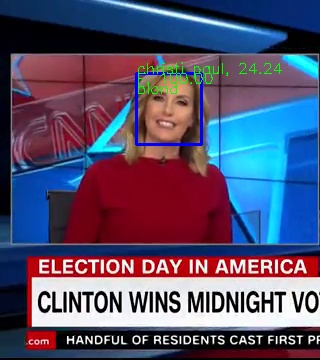}
    \caption{Frame 1}
  \end{subfigure}
  \begin{subfigure}{0.49\columnwidth}
    \includegraphics[width=0.98\columnwidth]{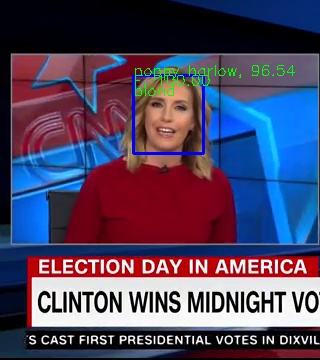}
    \caption{Frame 2}
  \end{subfigure}
  \caption{Two example frames from the same scene with an inconsistent attribute
  (the identity) from the TV news use case.}
  \label{fig:tv-news}
\end{figure}

\begin{figure}
  \begin{subfigure}{0.99\columnwidth}
    \includegraphics[width=0.98\columnwidth]{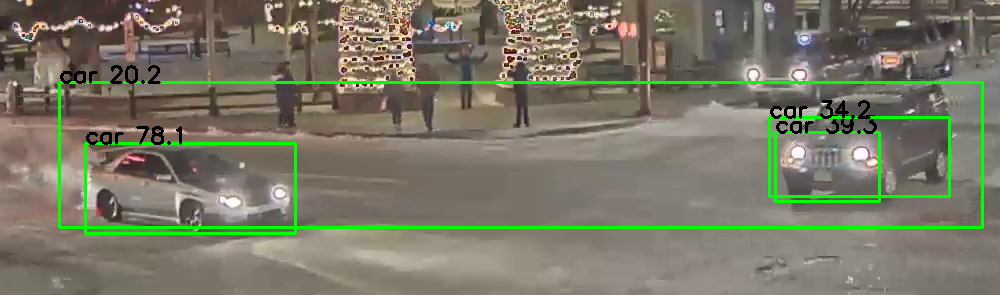}
    \caption{Example error 1.}
  \end{subfigure}
  \begin{subfigure}{0.99\columnwidth}
    \includegraphics[width=0.98\columnwidth]{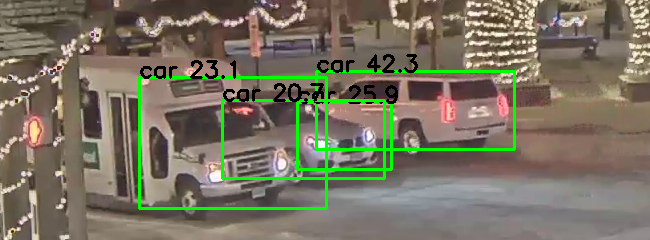}
    \caption{Example error 2.}
  \end{subfigure}
  \vspace{-0.5em}
  \caption{Examples errors when three boxes highly overlap (see
  \code{multibox} in Section~\ref{sec:eval}). Best viewed in color.}
  \label{fig:multibox}
\end{figure}

\begin{figure}
  \begin{subfigure}{0.99\columnwidth}
    \includegraphics[width=0.98\columnwidth]{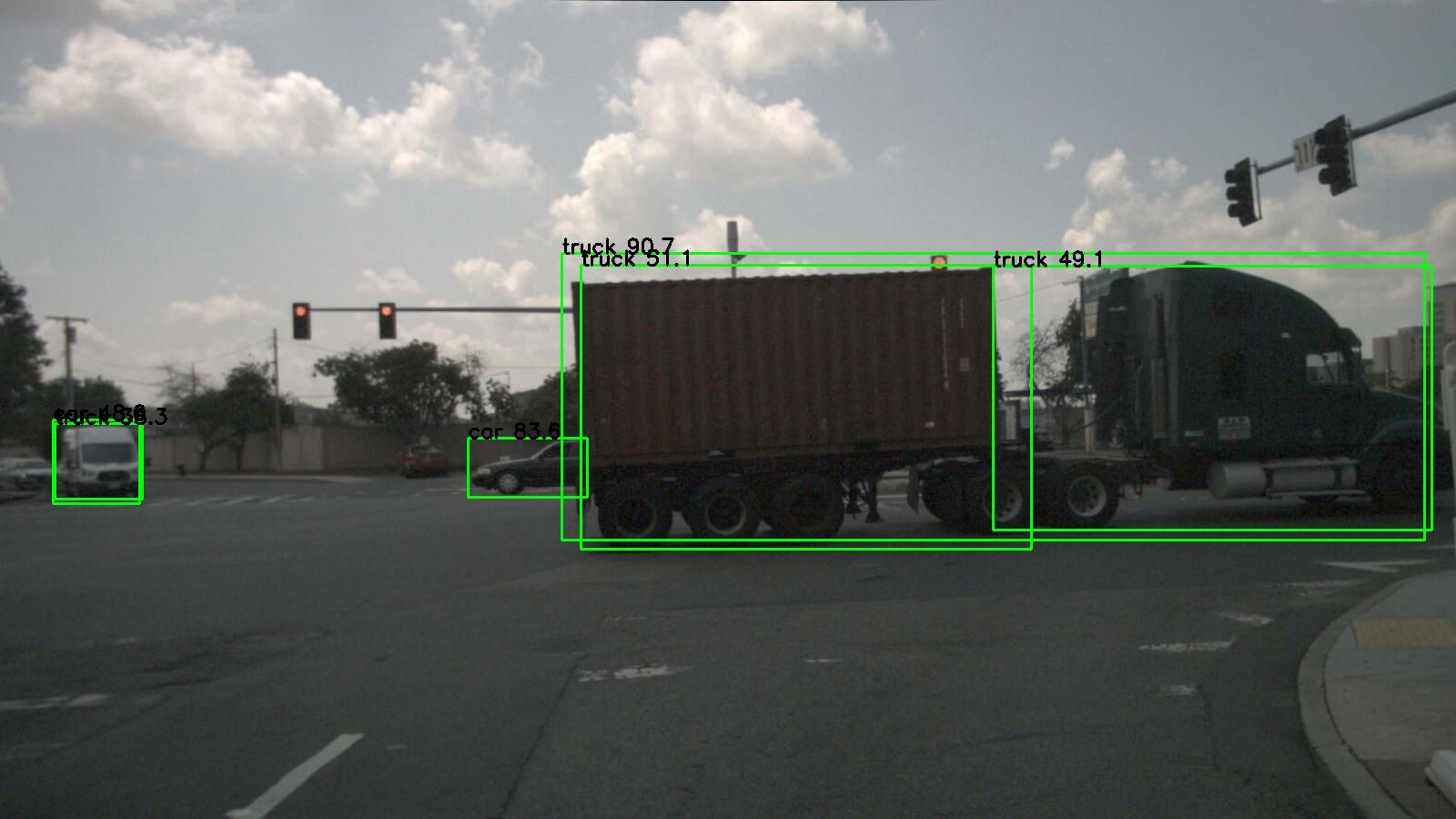}
    \caption{Example error flagged by \code{multibox}. SSD predicts three trucks
    when only one should be detected.}
  \end{subfigure}
  \begin{subfigure}{0.99\columnwidth}
    \includegraphics[width=0.98\columnwidth]{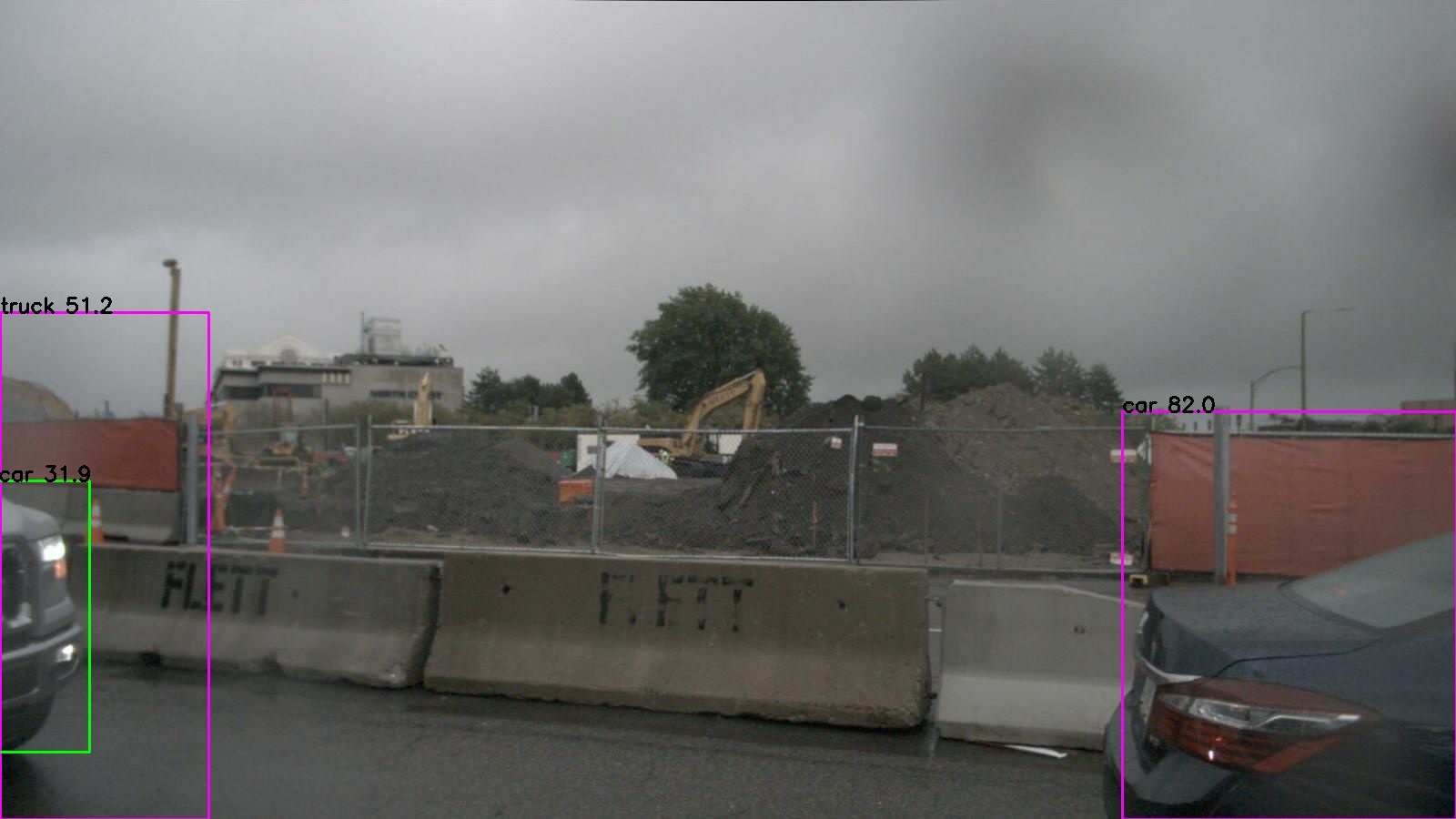}
    \caption{Example error flagged by \code{agree}. SSD misses the car on the
    right and the LIDAR model predicts the truck on the left to be too large.}
  \end{subfigure}
  \vspace{-0.5em}
  \caption{Examples of errors that the \code{multibox} and \code{agree}
  assertions catch for the NuScenes dataset. LIDAR model boxes are in pink and
  SSD boxes are in green. Best viewed in color.}
  \label{fig:nuscenes-examples}
\end{figure}

In this section, we illustrate several errors caught by the model assertions
used in our evaluation.

First, we show an example error in the TV news use case in
Figure~\ref{fig:tv-news}. Recall that these assertions were generated with our
consistency API (\S\ref{sec:synth}). In this example, the identifier is the
box's \code{sceneid} and the attribute is the \code{identity}.

Second, we show an example error for the visual analytics use case in
Figure~\ref{fig:multibox} for the \code{multibox} assertion. Here, SSD
erroneously detects multiple cars when there should be one.

Third, we show two example errors for the AV use case in
Figure~\ref{fig:nuscenes-examples} from the \code{multibox} and \code{agree}
assertions.

\section{Classes of Model Assertions}
\label{sec:ma-table}

\begin{table*}
\centering
\setlength\itemsep{5em}
\begin{tabularx}{\linewidth}{lllX}
  Assertion class & \specialcell{Assertion\\sub-class} & Description & Examples \\
  \hline \hline

  \specialcell{Consistency\\ \xspace}
    & \specialcell{Multi-source\\ \xspace}
        & \specialcell{Model outputs from multiple\\sources should agree}
        & \vspace{-2.25em} \begin{itemize}[leftmargin=*]
          \setlength\itemsep{0em}
          \item Verifying human labels (e.g., number of labelers that disagree)
          \item Multiple models (e.g., number of models that disagree)
        \end{itemize} \\
    & \specialcell{Multi-modal\\ \xspace}
        & \specialcell{Model outputs from multiple\\modes of data should agree}
        & \vspace{-2.25em} \begin{itemize}[leftmargin=*]
          \setlength\itemsep{0em}
          \item Multiple sensors (e.g., number of disagreements from LIDAR and
          camera models)
          \item Multiple data sources (e.g., text and images)
        \end{itemize}\\
    & \specialcell{Multi-view\\ \xspace}
        & \specialcell{Model outputs from multiple views\\of the same data should
        agree}
        & \vspace{-2.25em} \begin{itemize}[leftmargin=*]
          \setlength\itemsep{0em}
          \item Video analytics (e.g., results from overlapping views of
          different cameras should agree)
          \item Medical imaging (e.g., different angles should agree)
        \end{itemize}\\
  \hline

  \specialcell{Domain\\knowledge}
    & \specialcell{Physical\\ \xspace}
    & \specialcell{Physical constraints\\on model outputs}
        & \vspace{-2.25em} \begin{itemize}[leftmargin=*]
          \setlength\itemsep{0em}
          \item Video analytics (e.g., cars should not flicker)
          \item Earthquake detection (e.g., earthquakes should appear across
          sensors in physically consistent ways)
          \item Protein-protein interaction (e.g., number of overlapping atoms)
        \end{itemize}\\
    & \specialcell{Unlikely\\scenario}
    & \specialcell{Scenarios that are\\unlikely to occur}
        & \vspace{-2.25em} \begin{itemize}[leftmargin=*]
          \setlength\itemsep{0em}
          \item Video analytics (e.g., maximum confidence of 3 vehicles that highly overlap),
          \item Text generation (e.g., two of the same word should not appear sequentially)
        \end{itemize}
        \\
  \hline

  \specialcell{Perturbation\\ \xspace}
    & \specialcell{Insertion\\ \xspace}
        & \specialcell{Inserting certain types of data\\should not modify model
        outputs}
        & \vspace{-2.25em} \begin{itemize}[leftmargin=*]
          \setlength\itemsep{0em}
          \item Visual analytics (e.g., synthetically adding a car to a frame of video
        should be detected as a car),
          \item LIDAR detection (e.g., similar to visual analytics)
        \end{itemize} \\
    & \specialcell{Similar\\ \xspace \\ \xspace}
        & \specialcell{Replacing parts of the input with\\similar data should
        not modify\\model outputs}
        & \vspace{-3.0em} \begin{itemize}[leftmargin=*]
          \setlength\itemsep{0em}
          \item Sentiment analysis (e.g., classification should not change with
        synonyms)
          \item Object detection (e.g., painting objects different colors should
          not change the detection)
        \end{itemize} \\
    & \specialcell{Noise\\ \xspace}
        & \specialcell{Adding noise should not\\modify model outputs}
        & \vspace{-2.25em} \begin{itemize}[leftmargin=*]
          \setlength\itemsep{0em}
          \item Image classification (e.g., small Gaussian noise should not affect classification)
          \item Time series (e.g., small Gaussian noise should not affect
          time series classification)
        \end{itemize} \\
  \hline

  \specialcell{Input\\validation}
    & \specialcell{Schema\\validation}
        & \specialcell{Inputs should\\conform to a schema}
        & \vspace{-2.25em} \begin{itemize}[leftmargin=*]
          \setlength\itemsep{0em}
          \item Boolean features should not have inputs that are not 0 or 1
          \item All features should be present
        \end{itemize}
\end{tabularx}
\caption{Example of model assertions. We describe several assertion classes,
sub-classes, and concrete instantiations of each class. In parentheses, we
describe a potential severity score or an application.
}
\label{table:mas}
\end{table*}

We present a non-exhaustive list of common classes of model assertions in
Table~\ref{table:mas} and below. Namely, we describe how one might look for
assertions in other domains.

Our taxonomization is not exact and several examples will contain features from
several classes of model assertions. Prior work on schema
validation~\cite{polyzotis2019data, baylor2017tfx} and data
augmentation~\cite{wang2017effectiveness, taylor2017improving} can be cast in
the model assertion framework. As these have been studied, we do not focus on
these classes of assertions in this work.

\minihead{Consistency assertions}
An important class of model assertions checks the consistency across multiple
models or sources of data. The multiple sources of data could be the output of
multiple ML models on the same data, multiple sensors, or multiple views of the
same data. The output from the various sources should agree and consistency
model assertions specify this constraint. These assertions can be generated via
our API as described in \S\ref{sec:synth}.

\minihead{Domain knowledge assertions}
In many physical domains, domain experts can express physical constraints or
unlikely scenarios. As an example of a physical constraint, when predicting how
proteins will interact, atoms should not physically overlap. As an example of an
unlikely scenario, boxes of the visible part of cars should not highly overlap
(Figure~\ref{fig:multibox}). In particular, model assertions of unlikely
scenarios may not be 100\% precise, i.e., will be soft assertions.

\minihead{Perturbation assertions}
Many domains contain input and output pairs that can be perturbed (perhaps
jointly) such that the output does not change. These perturbations have been
widely studied through the lens of data
augmentation~\cite{wang2017effectiveness, taylor2017improving} and adversarial
examples~\cite{goodfellow2015explaining, athalye2018synthesizing}.

\minihead{Input validation assertions}
Domains that contain schemas for the input data can have model assertions that
validate the input data based on the schema~\cite{polyzotis2019data,
baylor2017tfx}. For example, boolean inputs that are encoded with integral
values (i.e., 0 or 1) should never be negative.  This class of assertions is an
instance of preconditions for ML models.

\section{Hyperparameters}
\label{sec:hyperparameters}

\minihead{Hyperparameters for active learning experiments}
For \code{night-street}, we used 300,000 frames of one day of video for the
training and unlabeled data. We sampled 100 frames per round for five rounds and
used 25,000 frames of a different day of video for the test set. Due to the cost
of obtaining labels, we ran each trial twice.

For the NuScenes dataset, we used 350 scenes to bootstrap the LIDAR model, 175
scenes for unlabeled/training data for SSD, and 75 scenes for validation (out of
the original 850 labeled scenes). We trained for one epoch at a learning rate of
$5 \times 10^{-5}$. We ran 8 trials.

For the ECG dataset, we train for 5 rounds of active learning with 100 samples
per round. We use a learning rate of 0.001 until the loss plateaus, which the
original training code did.

%
%


\begin{figure}
  \begin{subfigure}{0.99\columnwidth}
    \includegraphics[width=0.99\columnwidth]{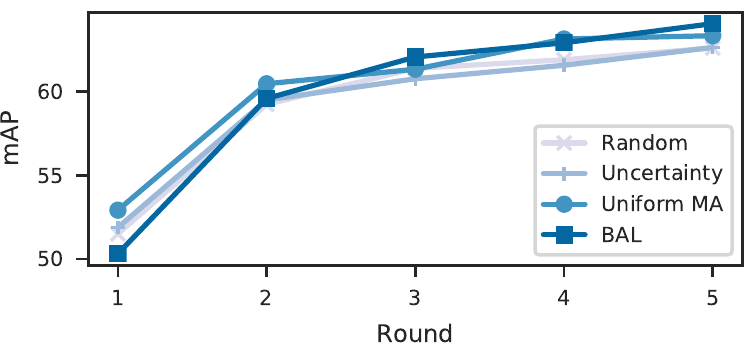}
    \vspace{-0.5em}
    \caption{Active learning for \texttt{night-street}.}
  \end{subfigure}
  \begin{subfigure}{0.99\columnwidth}
    \includegraphics[width=0.99\columnwidth]{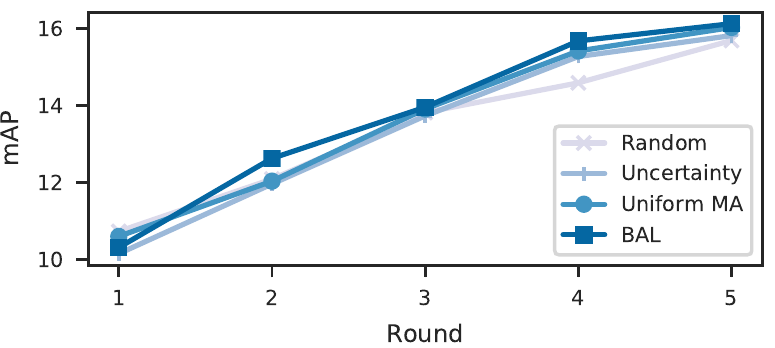}
    \vspace{-0.5em}
    \caption{Active learning for NuScenes.}
  \end{subfigure}
  \vspace{-0.5em}
  \caption{Performance of random sampling, uncertainty sampling, uniform
  sampling from model assertions, and BAL for
  active learning. The round is the round of data collection (see
  \S\ref{sec:alg}). As shown, BAL improves accuracy on unseen data and can
  achieve the same accuracy (62\% mAP) as random sampling with 40\% fewer labels
  for \texttt{night-street}. BAL also outperforms both baselines for the
  NuScenes dataset.}
  \vspace{-0.5em}
  \label{fig:multi-active-all}
\end{figure}

\section{Full Active Learning Figures}
\label{sec:full-al}

We show active learning results for all rounds in
Figure~\ref{fig:multi-active-all}.

\section{Model Assertions can Identify Errors in Human Labels}
\label{sec:ma-human-labels}

We further asked whether model assertions could be used to identify errors in
human-generated labels, i.e., a human is acting as a ``ML model.'' While
verification of human labels has been studied in the context of
crowd-sourcing~\cite{hirth2013analyzing, tran2013efficient}, several production
labeling services (e.g., Scale~\cite{scale}) do not provide annotator
identification which is necessary to perform this verification. We deployed a
model assertion in which we tracked objects across frames of a video using an
automated method and verified that the same object in different frames had the
same label.

\begin{table}
\centering
\small
\begin{tabular}{ll}
  Description   & Number \\ \hline
  All labels    & 469 \\
  Errors        & 32 \\
  Errors caught & 4
\end{tabular}
\vspace{-0.5em}
\caption{Number of labels, errors, and errors caught from model assertions for
Scale-annotated images for the video analytics task. As shown, model assertions
caught 12.5\% of the errors in this data.}
\vspace{-1em}
\label{table:scale}
\end{table}

We obtained labels for 1,000 random frames from \code{night-street} from Scale
AI~\cite{scale}, which is used by several autonomous vehicle companies.
Table~\ref{table:scale} summarizes our results. Scale returned 469 boxes, which we
manually verified for correctness. There were no localization errors, but there
were 32 classification errors, of which the model assertion caught 12.5\%. Thus,
we see that model assertions can also be used to verify human labels.


\end{document}